\documentclass[11pt]{article}
\usepackage[final]{acl}

\usepackage{times}
\usepackage{latexsym}
\usepackage[T1]{fontenc}
\usepackage[utf8]{inputenc}
\usepackage{microtype}
\usepackage{inconsolata}
\usepackage{amsmath}
\usepackage{amssymb}
\usepackage{graphicx}
\usepackage{booktabs}
\usepackage{multirow}
\usepackage{url}
\usepackage{xcolor}
\usepackage{tikz}
\usetikzlibrary{arrows.meta,positioning,fit,backgrounds,calc,decorations.pathreplacing}

\title{TextNCA: Neural Cellular Automata for Language Modeling\\via Hierarchical Local Attention}

\author{%
  Avni Mittal\textsuperscript{1} \quad
  Avinash Anand\textsuperscript{2} \quad
  Ashutosh Kumar\textsuperscript{3} \quad
  Dikshant Kukreja\textsuperscript{3} \\
  \textbf{Kritarth Prasad}\textsuperscript{3} \quad
  \textbf{Sushane Dulloo}\textsuperscript{3} \quad
  \textbf{Erik Cambria}\textsuperscript{4} \quad
  \textbf{Timothy Liu}\textsuperscript{5} \\
  \textbf{Zhengkui Wang}\textsuperscript{2} \quad
  \textbf{Rajiv Ratn Shah}\textsuperscript{3} \\[4pt]
  {\small \textsuperscript{1}Independent Researcher \quad
   \textsuperscript{2}Singapore Institute of Technology, Singapore} \\
  {\small \textsuperscript{3}IIIT Delhi, India \quad
   \textsuperscript{4}Nanyang Technological University, Singapore} \\
  {\small \textsuperscript{5}NVIDIA AI Technology Centre, Singapore}
}

\begin{document}
\maketitle

\begin{abstract}
Can a strictly local, iterated, weight-shared computation primitive support language modelling, and which of those three properties actually drives the model's behaviour? We define \textsc{TextNCA}, a 1D causal windowed-attention realisation of the Neural Cellular Automaton primitive, and study a hierarchical variant that cascades three stages with windows $w \in \{8, 32, 128\}$ and $T_s$ shared-weight iterations per stage, all on WikiText-103 at roughly 30M parameters and 60k training steps. The model does not match a parameter-matched Transformer at this scale (Hier-TextNCA $60.3$ vs.\ Transformer-6L $52.8$ and Transformer-12L $44.7$ PPL), so we treat it as an analytical probe rather than a proposed alternative. The behaviour we observe is largely explained by the staged narrow-to-wide schedule: a non-iterating sliding-window Transformer that reuses the same schedule comes within $+4.1$ PPL of the iterated model, while reversing, flattening, or breaking the monotonic ordering of the schedule costs between $+16.7$ and $+70.8$ PPL. Iteration adds a smaller bounded benefit on top of the schedule, with a clear optimum at $T_s{=}4$ and a U-shaped degradation beyond it. The GRU gate and learned per-step embeddings are required for that benefit to appear, and training with random $T_s$ yields an inference-time iteration-count knob at the cost of substantially higher absolute PPL. We position the work as a controlled reading of which parts of NCA-style computation carry the weight in language modelling.
\end{abstract}

\section{Introduction}
\label{sec:intro}

Neural Cellular Automata (NCAs) compute by applying a single shared local rule in place and iterating it until a global structure emerges \citep{mordvintsev2020growing,wolfram2002new}. They are characterised by three properties: strictly local perception, weight-shared iteration, and a gated in-place update. None of these is present in the dominant language modelling paradigm of stacked distinct transformer layers with global self-attention \citep{vaswani2017attention,kaplan2020scaling}, and NCAs themselves have been studied almost exclusively for images and textures \citep{mordvintsev2020growing,mordvintsev2021texture,palm2022variational}. This paper asks what each property does when the primitive is applied to language modelling.

Existing language models occupy different corners of this design space. Universal Transformers~\citep{dehghani2019universal} and Relaxed Recursive Transformers~\citep{bae2025relaxed} have iterated weight sharing but retain global attention and use a stateless residual update; ALBERT \citep{lan2020albert} shares parameters across layers without iteration; sliding-window transformers \citep{beltagy2020longformer,zaheer2020big} have local attention but neither weight sharing nor iteration. None combines all three NCA properties.

We introduce \textsc{TextNCA}, a 1D causal windowed-attention realisation of the NCA primitive, instantiated as a hierarchical model whose three stages cascade window sizes $w \in \{8, 32, 128\}$ with $T_s$ shared-weight iterations per stage. We treat the model as a probe rather than a performance bid: at matched training compute on WikiText-103, transformer baselines reach lower perplexity, and the contribution of this paper is analytical. We say a component is ``load-bearing'' when removing or perturbing it causes a large measured PPL increase relative to the flagship. Our contributions are:

\begin{enumerate}
\item \textbf{A controlled probe of NCA-inspired local iterative LMs.} We map the NCA primitive to a causal language model and run a seven-axis ablation: perception kernel, gate, step embedding, orchestration, stage count $K$, window schedule, and iteration count $T_s$.
\item \textbf{The narrow-to-wide staged context schedule is the dominant architectural factor.} Schedule perturbations cost tens of PPL, while a non-iterating stage-aware control comes within $+4.1$ PPL of the iterated flagship. The non-monotone schedule shows that \emph{starting narrow} matters independently of strict ordering.
\item \textbf{Iteration is a smaller bounded contributor that behaves as an effective-depth knob.} A from-scratch $T_s$ sweep is sharply U-shaped with an optimum at $T_s{=}4$; the benefit requires both the GRU gate and the learned per-step embeddings.
\item \textbf{Stochastic-iteration training enables an inference-time iteration-count knob, at substantial cost.} Training with $T_s \sim \mathcal{U}\{2,4,6\}$ and sinusoidal step embeddings yields a wide PPL valley where the deterministic counterpart diverges; the cost is substantially higher absolute PPL.
\end{enumerate}

\begin{figure*}[t]
\centering
\resizebox{\textwidth}{!}{%
\begin{tikzpicture}[
    font=\sffamily,
    >=Stealth,
    stage_red/.style={rectangle, draw, rounded corners=2pt, fill=red!15, align=center, minimum height=0.75cm, minimum width=0.95cm, thick, font=\sffamily\scriptsize},
    stage_blue/.style={rectangle, draw, rounded corners=2pt, fill=blue!15, align=center, minimum height=0.75cm, minimum width=0.95cm, thick, font=\sffamily\scriptsize},
    stage_green/.style={rectangle, draw, rounded corners=2pt, fill=green!15, align=center, minimum height=0.75cm, minimum width=0.95cm, thick, font=\sffamily\scriptsize},
    subcomp/.style={rectangle, draw, rounded corners=2pt, align=center, minimum height=0.65cm, minimum width=1.6cm, thick, font=\sffamily\scriptsize},
    attn_comp/.style={subcomp, fill=cyan!15},
    ln_comp/.style={subcomp, fill=gray!10, minimum height=0.45cm},
    ffn_comp/.style={subcomp, fill=yellow!20},
    gate_comp/.style={subcomp, fill=purple!18, rounded corners=6pt},
    io_box/.style={rectangle, draw, rounded corners=2pt, fill=orange!15, align=center, minimum height=0.75cm, minimum width=1.1cm, thick, font=\sffamily\scriptsize},
    io_node/.style={align=center, font=\sffamily\small\bfseries},
    arrow/.style={->, thick, >=Stealth},
    skip_arrow/.style={->, thick, dashed, draw=gray!70},
    cell_bg/.style={rectangle, draw=gray!50, rounded corners=4pt, fill=gray!3, inner sep=6pt, dashed}
]

    \node[io_node] (hin) {$\mathbf{h}_{t-1}$};

    \node[circle, draw, thick, inner sep=1pt, right=0.35cm of hin, fill=white, font=\scriptsize] (plus) {$+$};
    \node[font=\scriptsize, below=0.05cm of plus, text=gray!70] {$\mathbf{s}_t$};

    \node[attn_comp, right=0.4cm of plus]   (attn) {LocalAttn($w_k$)};
    \node[ln_comp,   right=0.3cm of attn]   (ln)   {LayerNorm};
    \node[ffn_comp,  right=0.3cm of ln]     (ffn)  {FFN};
    \node[gate_comp, right=0.35cm of ffn]   (gate) {GRU};
    \node[io_node,   right=0.4cm of gate]   (hout) {$\mathbf{h}_t$};

    \begin{scope}[on background layer]
      \node[cell_bg, fit=(plus)(attn)(ln)(ffn)(gate)] (cellbg) {};
    \end{scope}
    \node[font=\sffamily\footnotesize\bfseries, above right=0.35cm and 0pt of cellbg.north west, anchor=south west]
          {(a) NCA cell (applied $T_s$ times per stage)};

    \draw[arrow] (hin)  -- (plus);
    \draw[arrow] (plus) -- (attn);
    \draw[arrow] (attn) -- (ln);
    \draw[arrow] (ln)   -- (ffn);
    \draw[arrow] (ffn)  -- (gate);
    \draw[arrow] (gate) -- (hout);
    \draw[skip_arrow] (hin.north) |- ([yshift=0.45cm]attn.north) -| (gate.north);
    \node[font=\tiny, text=gray!70] at ([yshift=0.60cm]ffn.north) {residual skip};

    \node[font=\sffamily\tiny, below=0.1cm of gate, text=gray!80, align=center]
         {$z,r,\tilde{\mathbf{h}}$\\gated update};

    \node[io_box, below=2.3cm of hin.south, xshift=-0.1cm] (embed) {Embed};
    \node[io_node, left=0.35cm of embed, font=\sffamily\footnotesize] (xin) {Tokens};

    \foreach \i [evaluate=\i as \prev using int(\i-1)] in {1,...,4} {
      \ifnum\i=1
        \node[stage_red, right=0.45cm of embed] (s1-\i) {$w{=}8$\\\#\i};
      \else
        \node[stage_red, right=0.08cm of s1-\prev] (s1-\i) {$w{=}8$\\\#\i};
      \fi
    }
    \foreach \i [evaluate=\i as \prev using int(\i-1)] in {1,...,4} {
      \ifnum\i=1
        \node[stage_blue, right=0.35cm of s1-4] (s2-\i) {$w{=}32$\\\#\i};
      \else
        \node[stage_blue, right=0.08cm of s2-\prev] (s2-\i) {$w{=}32$\\\#\i};
      \fi
    }
    \foreach \i [evaluate=\i as \prev using int(\i-1)] in {1,...,4} {
      \ifnum\i=1
        \node[stage_green, right=0.35cm of s2-4] (s3-\i) {$w{=}128$\\\#\i};
      \else
        \node[stage_green, right=0.08cm of s3-\prev] (s3-\i) {$w{=}128$\\\#\i};
      \fi
    }
    \node[io_box, right=0.45cm of s3-4] (head) {LN +\\LM Head};
    \node[io_node, right=0.3cm of head, font=\sffamily\footnotesize] (logits) {Logits};

    \draw[arrow] (xin)   -- (embed);
    \draw[arrow] (embed) -- (s1-1);
    \foreach \a/\b in {s1-1/s1-2, s1-2/s1-3, s1-3/s1-4, s2-1/s2-2, s2-2/s2-3, s2-3/s2-4, s3-1/s3-2, s3-2/s3-3, s3-3/s3-4} {
      \draw[arrow] (\a) -- (\b);
    }
    \draw[arrow, red!60,  line width=1.3pt] (s1-4) -- (s2-1);
    \draw[arrow, blue!60, line width=1.3pt] (s2-4) -- (s3-1);
    \draw[arrow] (s3-4) -- (head);
    \draw[arrow] (head)  -- (logits);

    \draw[decorate, decoration={brace, amplitude=4pt, mirror}, thick, red!70]
      ([yshift=-0.18cm]s1-1.south west) -- ([yshift=-0.18cm]s1-4.south east)
      node[midway, below=3pt, font=\sffamily\scriptsize\bfseries, red!75] (b1) {Stage 1 $\cdot$ $\theta_1$};
    \node[font=\sffamily\scriptsize\itshape, red!70, below=1pt of b1] {$w{=}8$ (narrow)};

    \draw[decorate, decoration={brace, amplitude=4pt, mirror}, thick, blue!70]
      ([yshift=-0.18cm]s2-1.south west) -- ([yshift=-0.18cm]s2-4.south east)
      node[midway, below=3pt, font=\sffamily\scriptsize\bfseries, blue!75] (b2) {Stage 2 $\cdot$ $\theta_2$};
    \node[font=\sffamily\scriptsize\itshape, blue!70, below=1pt of b2] {$w{=}32$ (mid)};

    \draw[decorate, decoration={brace, amplitude=4pt, mirror}, thick, green!55!black]
      ([yshift=-0.18cm]s3-1.south west) -- ([yshift=-0.18cm]s3-4.south east)
      node[midway, below=3pt, font=\sffamily\scriptsize\bfseries, green!45!black] (b3) {Stage 3 $\cdot$ $\theta_3$};
    \node[font=\sffamily\scriptsize\itshape, green!45!black, below=1pt of b3] {$w{=}128$ (wide)};

    \node[font=\sffamily\footnotesize\bfseries, above right=0.05cm and -0.1cm of embed.north west, anchor=south west]
         {\textbf{(b) Staged pipeline} --- 3 stages $\times$ $T_s{=}4$ iterations $=$ 12 NCA steps total};

    \begin{scope}[shift={([xshift=1.4cm, yshift=-0.7cm]hout.east)}]
      \node[font=\sffamily\footnotesize\bfseries, anchor=south west] at (-0.1, 1.95)
           {(c) Effective receptive field at query};
      \foreach \row in {1.3, 0.65, 0.0} {
        \foreach \c in {0,...,14} {
          \fill[gray!15, draw=gray!50]
            ({\c*0.22}, \row) rectangle ({\c*0.22+0.2}, \row+0.38);
        }
      }
      \foreach \c in {4,...,11} {
        \fill[red!45, draw=red!75]
          ({\c*0.22}, 1.3) rectangle ({\c*0.22+0.2}, 1.68);
      }
      \foreach \c in {0,...,11} {
        \fill[blue!40, draw=blue!75]
          ({\c*0.22}, 0.65) rectangle ({\c*0.22+0.2}, 1.03);
      }
      \foreach \c in {0,...,14} {
        \fill[green!40, draw=green!70]
          ({\c*0.22}, 0.0) rectangle ({\c*0.22+0.2}, 0.38);
      }
      \draw[->, thick] (2.52, -0.35) -- (2.52, -0.05);
      \node[font=\sffamily\tiny, below=-2pt of {(2.52, -0.35)}] {query position};
      \node[font=\sffamily\tiny, red!80,         anchor=east] at (-0.03, 1.49) {$w{=}8$};
      \node[font=\sffamily\tiny, blue!80,        anchor=east] at (-0.03, 0.84) {$w{=}32$};
      \node[font=\sffamily\tiny, green!45!black, anchor=east] at (-0.03, 0.19) {$w{=}128$};
    \end{scope}

\end{tikzpicture}%
}
\caption{\textbf{Hierarchical TextNCA (staged) architecture.}
\textbf{(a)} A single NCA cell: local attention with window $w_k$, LayerNorm, FFN, and a GRU gate that combines the previous state $\mathbf{h}_{t-1}$ (dashed residual) with the transformed signal. A learned step embedding $\mathbf{s}_t$ is added at the input so the shared weights can still specialise across iterations.
\textbf{(b)} The staged pipeline. Three stages with windows $w \in \{8, 32, 128\}$ iterate $T_s{=}4$ times each (12 NCA steps total). Each stage has its own parameter set $\theta_k$ but shares these weights across its four iterations. Bold red/blue arrows mark the two-stage transitions, where the receptive field expands to the next scale.
\textbf{(c)} Effective receptive field for a query near the end of a 15-token window. Stage 1 sees 8 nearby tokens (red); Stage 2 covers the local context (blue); Stage 3 reaches the full sequence (green). The receptive-field structure motivates the per-step loss decomposition: stage 3 is observed to carry essentially all of the next-token loss reduction.}
\label{fig:architecture}
\end{figure*}
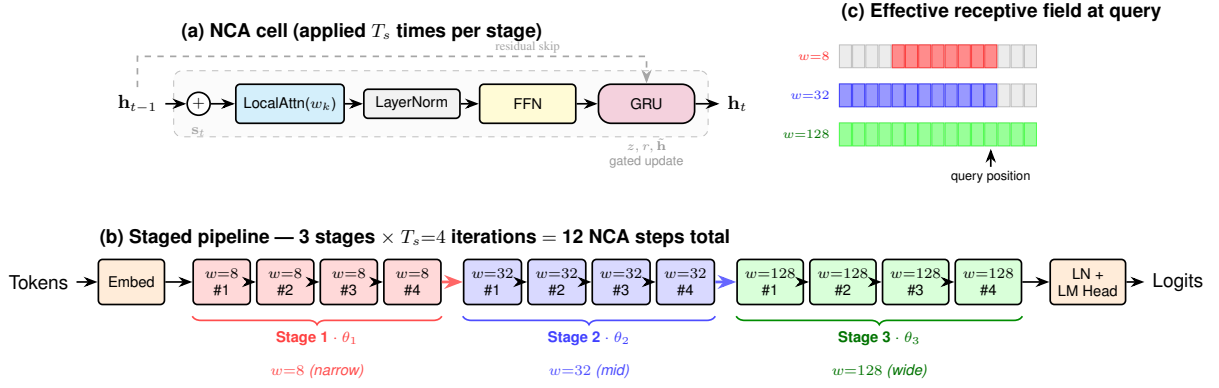

\section{Related Work}
\label{sec:related_work}

\textbf{Iterated and weight-shared transformers.} Universal Transformers \citep{dehghani2019universal} apply a shared block for $T$ iterations with \emph{global} attention and a stateless residual update; under our protocol, UT reaches PPL 91.7 vs.\ Hier-TextNCA's 60.3, indicating weight sharing under global attention is not what drives Hier-TextNCA. ALBERT \citep{lan2020albert} shares parameters across layers without iteration; Relaxed Recursive Transformers \citep{bae2025relaxed} share layer blocks with low-rank relaxations and retain global attention; looped-transformer work \citep{geiping2025latent,saunshi2025reasoning} explores test-time iteration depth at much larger scales. Our $T_s$ U-shape at 30M is independent evidence for the reading that iteration in LMs behaves as effective depth, and not as a refinement axis. Earlier, deeply iterative architectures (Neural GPU \citep{kaiser2016neural}, DEQ \citep{bai2019deep}) do not target autoregressive LM. \\
\newline
\textbf{Local, sparse, and efficient attention.} Longformer \citep{beltagy2020longformer} and BigBird \citep{zaheer2020big} combine sliding-window attention with sparse global tokens; sliding-window transformers without iteration or weight sharing are the family of our SWin-TF-Staged control. State-space and long-convolution models (Mamba \citep{gu2024mamba}, Hyena \citep{poli2023hyena}) replace attention entirely; comparison is a known gap. We tested GLA \citep{yang2024gated} as a perception kernel in the hierarchical body and found that despite GLA beating local softmax at single scale (129 vs.\ 140 PPL at 60k steps), the hierarchical body inverts this ordering (72.0 vs.\ 60.3), a non-trivial kernel--schedule interaction.\\
\newline
\textbf{NCA and position of this work.} NCAs were introduced for image generation \citep{mordvintsev2020growing} and extended to texture synthesis \citep{mordvintsev2021texture,palm2022variational}; subsequent work introduced hierarchical NCA variants for medical image segmentation, using multi-scale rules to propagate global information across the image \citep{kalkhof2023med,mittal2025medsegdiffnca}. Concurrent work uses NCA-generated data to pre-train transformers \citep{lee2026ncadata}. TextNCA is the first work, to our knowledge, to use NCA-style computation as the language-model architecture. None of the prior architectures above combines all three NCA properties simultaneously; this paper is not novel because it uses iteration, locality, or weight sharing alone, but because it tests their conjunction under a hierarchical staged schedule and isolates which part actually drives the empirical behaviour.

\section{TextNCA: Architecture}
\label{sec:architecture}

\textsc{TextNCA} implements the NCA primitive as repeated weight-shared updates to a sequence of token states; Table~\ref{tab:nca-mapping} maps the image-domain NCA components onto our realisation. Each of the three stages applies the same local-attention update $T_s$ times under its own parameter set, so the model is a hierarchical NCA. TextNCA is, to our knowledge, the first language model to combine strictly local attention, iterated weight sharing, and a GRU-style gated in-place update. We use ``NCA'' as a structural description of the per-stage rule and treat the schedule of window sizes across stages as the architectural choice that carries the empirical work. Appendix~\ref{app:nca_primitive} situates the primitive against the image-domain NCA of \citet{mordvintsev2020growing}, and Appendix~\ref{app:arch_full} gives the full architecture specification.

\begin{table}[h]
\centering
\small
\setlength{\tabcolsep}{4pt}
\begin{tabular}{p{0.42\columnwidth} p{0.47\columnwidth}}
\toprule
\textbf{NCA property} & \textbf{TextNCA realisation} \\
\midrule
Cell state vector & Token hidden state $\mathbf{h}_i \in \mathbb{R}^d$ \\
Local perception kernel & Causal windowed attention, window $w$ \\
Shared update rule & Tied weights within each stage \\
Iterative in-place update & $\mathbf{h} \leftarrow \text{Gate}(\mathbf{h}, f(\mathbf{h}))$, $T_s$ times \\
Gated residual (``fire rate'') & GRU update/reset gates \\
Multi-scale extension & Stage cascade $w \in \{8, 32, 128\}$ \\
\bottomrule
\end{tabular}
\caption{NCA primitives mapped to \textsc{TextNCA}. The load-bearing NCA conjunction (locality, iterated weight sharing, gated in-place update) holds inside each stage; the full model is a hierarchical NCA with rule piecewise-constant across stages.}
\label{tab:nca-mapping}
\end{table}

\subsection{Core update}
\label{sec:core}

Given input tokens $\mathbf{x} = (x_1,\ldots,x_L)$, we initialize token states as
{\small
\begin{equation}
\mathbf{h}_0 =
\mathrm{TokenEmbed}(\mathbf{x}) + \mathrm{PosEmbed}(\mathbf{x})
\in \mathbb{R}^{B \times L \times d}
\end{equation}
}
The model then applies the same recurrent update for $T$ iterations:
\begin{equation}
\resizebox{0.95\columnwidth}{!}{$\displaystyle
\mathbf{h}_t =
\mathrm{Gate}\!\left(
\mathbf{h}_{t-1},
\mathrm{FFN}\!\left(
\mathrm{LN}\!\left(
\mathrm{Perceive}(\mathbf{h}_{t-1} + \mathbf{s}_t)
\right)
\right)
\right)
$}
\label{eq:step}
\end{equation}
where $\mathbf{s}_t \in \mathbb{R}^{d}$ is a learnable step embedding. The perception module, FFN, gate, and normalization parameters are \emph{shared across all iterations}. Final predictions are produced by
\begin{equation}
\mathrm{logits} =
\mathrm{LMHead}\!\left(\mathrm{LN}(\mathbf{h}_T)\right).
\end{equation}

In all main experiments, $\mathrm{Perceive}(\cdot)$ is local causal softmax attention with window size $w$, $\mathrm{Attn}(Q_i,K,V)$:
\begin{equation}
\mathrm{softmax}\!\left(
\frac{Q_i K_{[i-w+1:i]}^\top}{\sqrt{d_h}}
\right)
V_{[i-w+1:i]}
\end{equation}
The alternative perception kernels are used only for the ablation in Appendix~\ref{app:kernels}: local linear attention with a Taylor-2 feature map
$\phi(\mathbf{x}) {=} [1,\mathbf{x},\mathbf{x}\otimes\mathbf{x}/\sqrt{2}]$,
gated linear attention following \citet{yang2024gated}, and causal depthwise convolution with kernel size $w$. The main gate is a GRU-style update that combines the perception output $\mathbf{u}$ with the previous state $\mathbf{h}_{t-1}$:
\begin{align}
\mathbf{z} &=
\sigma(\mathbf{W}_z[\mathbf{h}_{t-1};\mathbf{u}]),
\\
\mathbf{r} &=
\sigma(\mathbf{W}_r[\mathbf{h}_{t-1};\mathbf{u}]), \\
\tilde{\mathbf{h}} &=
\tanh(\mathbf{W}_h[\mathbf{r}\odot\mathbf{h}_{t-1};\mathbf{u}]),
\\
\mathbf{h}_t &=
(1-\mathbf{z})\odot\mathbf{h}_{t-1}
+
\mathbf{z}\odot\tilde{\mathbf{h}}.
\end{align}
The gating ablation in Appendix~\ref{app:gates} replaces this GRU gate with sigmoid, highway, residual, or overwrite updates; this ablation yields the $+16.3$ PPL GRU-vs-residual penalty reported in \S\ref{sec:results}.

\subsection{Hierarchical staged TextNCA}
\label{sec:hierarchical}

The hierarchical variant processes tokens through $K$ sequential stages with strictly increasing window sizes. Each stage $k$ has its own parameter set $\theta_k= (\text{Perceive}_{w_k}, \text{LN}_k, \text{FFN}_k, \text{Gate}_k)$ and iterates $T_s$ times with $\theta_k$ held fixed. The output of stage $k$ becomes the input to stage $k{+}1$ with no reset. The window schedule is $[w_1,w_2,w_3] = [8,32,128]$.
For each stage $k = \{1..K\}$, the hidden state is refined $T_s$ times:
{\small
\begin{align}
\text{For } t &= 1..T_s: \nonumber \\
\mathbf{u} &= \text{FFN}_{\theta_k}(\text{LN}_{\theta_k}(\text{Perceive}_{\theta_k, w_k}(\mathbf{h} + \mathbf{s}_{(k-1)T_s + t}))) \\
\mathbf{h} &\leftarrow \text{Gate}_{\theta_k}(\mathbf{h}, \mathbf{u}).
\end{align}
}

We fix $K=3$ and $T_s=4$ (so total NCA steps $= 12$) based on the $T_s$ sweep in \S\ref{sec:iteration_ablation} (U-shaped with minimum at $T_s{=}4$) and the stage-count ablation in \S\ref{sec:schedule} ($K{=}3$ beats $K{=}2$ and $K{=}4$ at matched NCA steps).\\
\newline
\textbf{Receptive-field profile.} For causal local attention with window $w_k$ applied $T_s$ times in stage $k$, the receptive field at position $i$ grows additively: $R_K {=} 1 + \sum_{k=1}^{K} T_s(w_k - 1)$. For $T_s{=}4$ and $w \in \{8, 32, 128\}$, this gives $R_1{=}29, R_2{=}153, R_3{=}661$, a strict narrow-to-wide cascade, exceeding the 512-token training context only at stage 3. The dependencies separated by more than 152 tokens, therefore, cannot be resolved before stage 3 - the mechanistic content of the schedule choice, and \S\ref{sec:interp} reports the empirical consequence.\\
\newline
\textbf{Staging vs.\ interleaving.} Whether to expose the three window sizes \emph{as stages} (consecutive iterations at $w{=}8$, then $w{=}32$, then $w{=}128$) versus \emph{within each iteration} (cycle $[w{=}8 \to w{=}32 \to w{=}128 \to \text{Gate}]$ each step) makes a large difference. The staged design reaches 60.3 PPL while the interleaved variant reaches 94.4 PPL ($\Delta = 34.1$, Table~\ref{tab:orchestration}). This is a separate piece of evidence, alongside the schedule-perturbation ablations, that the \emph{ordering} of window sizes across iterations, not merely the set of windows used, is the load-bearing property.


%

\section{Experimental Setup}
\label{sec:experimental_setup}

\subsection{Training protocol and compute}
\label{sec:training}
We use full backpropagation through all $T$ steps with gradient checkpointing \citep{chen2016training} and AdamW (learning rate $6{\times}10^{-4}$, $\beta=(0.9,0.95)$, weight decay $0.1$, gradient clipping at $1.0$), with a cosine schedule and $2000$ warmup steps in bfloat16. All main-comparison models train for matched $60$k steps at an effective batch of $65{,}536$ tokens per step ($\approx 4$B tokens total) on a $500$M-token subset of SlimPajama \citep{cerebras2023slimpajama}, tokenised with the Mistral 32K BPE vocabulary at sequence length $512$. Hier-TextNCA runs at roughly $2{\times}$ the FLOPs and $2.7{\times}$ lower throughput than Transformer-6L per training step, because the 12 shared updates cannot be fused like independent layers; we therefore report both matched-parameter and matched-compute comparisons (\S\ref{sec:main_lm}, Appendix~\ref{app:flops_matched}). Per-run FLOPs, throughput and GPU-hour figures are in Appendix~\ref{app:compute}.

\subsection{Language-modelling evaluation}
\label{sec:lm_eval}
We evaluate on WikiText-103 \citep{merity2017pointer} and report token-level perplexity on its test split, with no further fine-tuning beyond LM pre-training. Because training uses SlimPajama with a Mistral 32K BPE vocabulary rather than the in-domain WikiText-103 vocabulary, our perplexities are not comparable to literature numbers that use adaptive softmax on in-domain data; we compare TextNCA only against transformer baselines trained under the same SlimPajama--Mistral protocol.

\subsection{Downstream tasks}
\label{sec:downstream_setup}
To test whether language-modelling trends transfer, we fine-tune each 60k-step checkpoint on three classification benchmarks and one extractive QA benchmark.

\paragraph{Datasets and splits.}
IMDb \citep{maas2011imdb}: binary sentiment, $25$k train / $25$k test. AG~News \citep{zhang2015character}: 4-class topic, $120$k train / $7.6$k test. SST-2 \citep{socher2013recursive}: binary sentiment, $\approx 67$k train / $872$ dev. SQuAD~v1.1 \citep{rajpurkar2016squad}: extractive QA, $\approx 87$k train / $10.6$k dev. For SST-2 and SQuAD~v1.1, whose test labels are not public, we evaluate on the official dev split following standard practice.

\paragraph{Fine-tuning.}
We fine-tune each pre-trained checkpoint on each task with AdamW for $5$ epochs (classification) or $3$ epochs (SQuAD), reusing the LM sequence length and tokeniser. Full hyperparameters (learning rates, batch sizes, head architecture) are in Appendix~\ref{app:reproducibility}.

\paragraph{Metrics and seeds.}
Classification: top-1 accuracy on the held-out evaluation split, reported as a mean over three seeds ($42, 1337, 2024$); per-seed standard deviations are in Appendix~\ref{app:downstream_full}. SQuAD: token-level F1 and exact-match on the dev split, computed with the official evaluator; single-seed.

\subsection{Baselines}
\label{sec:baselines}
All baselines are trained under the same protocol. \textbf{Transformer-6L} and \textbf{Transformer-12L} are standard-architecture references at two depths. \textbf{Universal Transformer} \citep{dehghani2019universal} isolates weight sharing under global attention. \textbf{SWin-TF-Staged} reuses Hier-TextNCA's narrow-to-wide schedule as layer windows while removing iteration and weight sharing, isolating the schedule from the other two NCA properties. \textbf{SWin-TF-6L} controls for window-locality without the staged schedule. Full configurations are in Appendix~\ref{app:arch_full}, Table~\ref{tab:configs_full}.

\section{Results}
\label{sec:results}
We report language-modelling perplexity on WikiText-103 (\S\ref{sec:schedule}--\S\ref{sec:component_ablations}) and downstream transfer to classification and extractive QA (\S\ref{sec:downstream}); the experimental protocol is described in \S\ref{sec:experimental_setup}.
\subsection{Main Finding: Schedule and Iteration}
\label{sec:schedule}
\begin{table}[t]
\centering
\small
\setlength{\tabcolsep}{4pt}
\begin{tabular}{lrr}
\toprule
\textbf{Variant} & \textbf{Params} & \textbf{PPL} \\
\midrule
\textbf{Staged $[8,32,128]{\times}T_s{=}4$ (flagship)} & 30.8M & \textbf{60.3} \\
Interleaved $[8,32,128]{\times}T{=}4$ & 27.7M & 94.4 \\
\midrule
\multicolumn{3}{l}{\textit{Window-schedule axis:}} \\
\midrule
Reversed $[128,32,8]$           & 30.8M & 131.1 \\
Uniform $[32,32,32]$            & 30.8M & 119.7 \\
Non-monotone $[8,128,32]$       & 30.8M & 77.0 \\
\midrule
\multicolumn{3}{l}{\textit{Stage-count axis (NCA steps $\approx 12$):}} \\
\midrule
$K{=}2$ $[16,128]{\times}T_s{=}6$         & 26.1M & 75.8 \\
$K{=}4$ $[4,16,64,128]{\times}T_s{=}3$    & 35.6M & 71.8 \\
\midrule
\multicolumn{3}{l}{\textit{Stage-aware non-iterating controls:}} \\
\midrule
\textbf{SWin-TF-Staged $[8,32,128]$ (6L)}   & 35.5M & \textbf{64.5} \\
SWin-TF-6L (uniform $w{=}128$)              & 35.5M & 63.2 \\
\bottomrule
\end{tabular}
\caption{Schedule and orchestration ablations at matched architecture and compute, 60k training steps, total NCA steps $\approx 12$. The interleaved variant uses a \emph{single} cell whose window is cycled per step (27.7M), versus the staged variant's three per-stage cells with distinct weights (30.8M); the 3.1M gap is the cost of going from one shared cell to three. Matching is on training compute, total NCA-step applications, and tokenizer / data / optimizer, not on parameter count.}
\label{tab:orchestration}
\end{table}

\begin{figure}[t]
\centering
\includegraphics[width=\columnwidth]{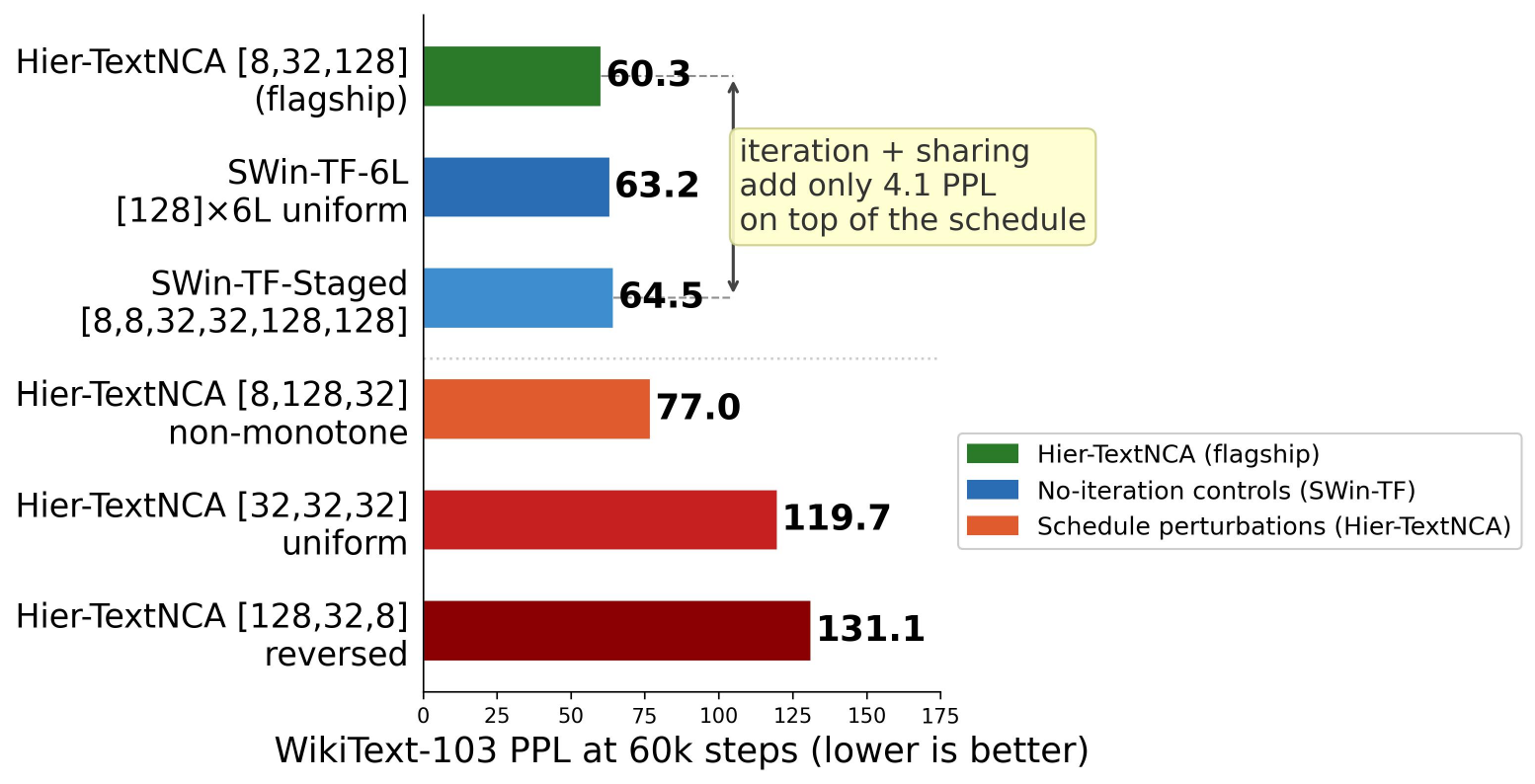}
\caption{The bracket marks the $+4.1$ PPL gap between Hier-TextNCA (60.3) and SWin-TF-Staged (64.5), against which the $+59.4$ to $+70.8$ PPL penalties from schedule perturbations should be read.}
\label{fig:schedule_bar}
\end{figure}
Table~\ref{tab:orchestration} reports the schedule and orchestration ablations; Figure~\ref{fig:schedule_bar} visualizes the headline comparison, the effect decomposed along three independent axes:\\
\newline
\textbf{(1) Window-schedule direction and shape.} Varying only the schedule produces forward $[8,32,128]$ at 60.3 PPL; reversed $[128,32,8]$ at 131.1 ($+70.8$); uniform $[32,32,32]$ at 119.7 ($+59.4$); non-monotone $[8,128,32]$ at 77.0 ($+16.7$). All three controls confirm that narrow-to-wide staged expansion is essential. The non-monotone variant retains the narrow first stage but inserts a wide middle stage, and its penalty is far smaller than reversing or flattening: \emph{starting narrow} matters independently of strict monotone ordering. Strict monotonicity is, therefore, the best instantiation of the more general principle of beginning narrow and expanding gradually.\\
\newline
\textbf{(2) Stage count.} At matched total NCA steps: $K{=}2$ ([16,128] with $T_s{=}6$, 75.8 PPL) and $K{=}4$ ([4,16,64,128] with $T_s{=}3$, 71.8 PPL) both lose meaningfully to $K{=}3$. Each stage needs at least $T_s{=}4$ steps to reach intra-stage convergence before the next scale begins; below that, the schedule's mechanistic benefit is incomplete.\\
\newline
\textbf{(3) Stage-aware Non-iterating Control.} A 6-layer transformer with no iteration and no weight sharing, whose layer windows are scheduled $[8,8,32,32,128,128]$ to match Hier-TextNCA's receptive field, reaches 64.5 PPL, only $+4.1$ behind Hier-TextNCA. Combined with the schedule perturbations, this gives our central reading: \emph{the narrow-to-wide staged window schedule accounts for the bulk of the architecture's effect; iterated weight sharing is a secondary contributor adding $\sim 4$ PPL on top.} The uniform-$w{=}128$ SWin-TF-6L control at 63.2 PPL is slightly better than SWin-TF-Staged in the non-iterating setting. We read this as evidence that the schedule's full benefit is conditional on the iterated weight-shared body (UT alone is 91.7 PPL, iteration with global attention is far worse than non-iterated transformers with appropriate windowing).
\subsection{Comparison to Transformers}
\label{sec:main_lm}

\begin{table}[t]
\centering
\small
\setlength{\tabcolsep}{4pt}
\begin{tabular}{lrrr}
\toprule
\textbf{Model} & \textbf{Params} & \textbf{tok/s} & \textbf{PPL} \\
\midrule
Transformer-12L                 & 54.5M & 103K & 44.7 \\
Transformer-6L                  & 35.5M & 300K & 52.8 \\
\textbf{Hier-TextNCA}           & \textbf{30.8M} & \textbf{112K} & \textbf{60.3} \\
SWin-TF-Staged $[8,32,128]$     & 35.5M & 295K & 64.5 \\
SWin-TF-6L ($w{=}128$)          & 35.5M & 298K & 63.2 \\
Universal Transformer           & 21.4M & 151K & 91.7 \\
Hier-TextNCA (Residual gate)    & 26.1M & 135K & 76.6 \\
Hier-TextNCA (GLA kernel)       & 30.9M &  65K & 72.0 \\
\bottomrule
\end{tabular}
\caption{WikiText-103 PPL at matched 60k steps with training-time tokens-per-second (per-GPU steady-state median on a Blackwell-6000, measured from training logs). Hier-TextNCA reaches 60.3 PPL; Transformer-6L (52.8) and Transformer-12L (44.7) do better. The contribution is analytical, not performance-based.}
\label{tab:main}
\end{table}

Training-step trajectories for the same runs are in Appendix~\ref{app:training_dynamics} (Figure~\ref{fig:eval_ppl}).

The matched-step comparison is the most favourable framing for TextNCA; under matched compute, the gap widens substantially. Hier-TextNCA also runs at $1.9\times$ the FLOPs and $2.7\times$ lower throughput than Transformer-6L per training step (Table~\ref{tab:main}). At FLOP-matched compute, Hier-TextNCA reaches PPL 70.6 versus Transformer-6L's 52.8; at wall-clock-matched compute, 80.1 vs.\ 52.8 (Appendix~\ref{app:flops_matched}, Figure~\ref{fig:pareto}). TextNCA is not computationally Pareto-superior to transformers at this scale; any ``parameter efficiency'' reading should be restricted to matched-parameter, matched-step protocols.\\
\newline
\textbf{Additional comparisons.}
The Universal Transformer baseline (21.4M parameters, global attention, weight sharing, stateless residual update) reaches 91.7 PPL, substantially worse than Hier-TextNCA and worse than every non-iterating sliding-window variant. Weight sharing under global attention is therefore not the architectural ingredient driving Hier-TextNCA's behaviour; the staged local schedule is. Substituting the GLA perception kernel into the hierarchical body degrades PPL from 60.3 to 72.0 ($+11.7$), despite GLA outperforming local softmax in the single-scale setting (129 vs.\ 140 PPL at 60k steps). We attribute this to a window-kernel interaction (§\ref{sec:component_ablations}, Figure~\ref{fig:kernel_ablation}).

\subsection{Ablations of Architectural Components}
\label{sec:component_ablations}

The schedule identifies \emph{which} architectural choice is load-bearing; component ablations identify \emph{which mechanisms} let the iterated variant realize its $+4.1$ PPL share over SWin-TF-Staged.\\
\newline
\textbf{Gating.} Replacing the GRU with a plain residual additive update (UT-style) drops PPL from 60.3 to 76.6 ($+16.3$), isolating the GRU as the load-bearing primitive that distinguishes TextNCA from a residual-only iterated transformer. Pure sigmoid gating diverges; SigmoidDropout stabilizes but underperforms (71.7). More details in Appendix~\ref{app:gates}, and Table~\ref{tab:gating_full}.\\
\newline
\textbf{Step embeddings.} Removing the learned per-iteration embeddings $\mathbf{s}_t$ inflates PPL to 82.7; replacing them with sinusoidal collapses PPL to 119.5 (Table~\ref{tab:step_embedding_full}). We read this jointly as evidence that \emph{some} parametric per-step conditioning is required for iteration to help, and that the sinusoidal choice in particular interacts poorly with deterministic short-horizon $T_s$; it does \emph{not} isolate learned embeddings against other parametric per-step conditioners (e.g.\ Adaptive Computation Time halting \citep{graves2016adaptive}, learned-frequency sinusoids, or FiLM-style step modulation), which we flag as future work.\\
\newline
\textbf{Iteration count $T_s$.}
\label{sec:iteration_ablation}
We retrain Hier-TextNCA from scratch at $T_s\in\{2,4,6,8\}$ with matched everything else; total NCA steps are $3T_s$. PPL is 80.3, 60.3, 73.9, 117.9 - sharply U-shaped with a minimum at $T_s{=}4$. Three readings: (i) $T_s{=}8$ (24 steps) is \emph{worse} than $T_s{=}2$ (6 steps) by 38 PPL, ruling out an iterative-refinement reading. (ii) Parameter count is essentially fixed ($\Delta<0.2\%$), so the U-shape reflects optimisation difficulty, not capacity. (iii) Direct dynamical evidence: $\|\Delta\mathbf{h}\|\approx 4$ at the trained horizon (Appendix~\ref{app:hidden_dynamics}); a contraction-mapping reading would predict $\|\Delta\mathbf{h}\|\to 0$. Together with the test-time analysis in \S\ref{sec:ttc}, $T_s$ behaves as an effective-depth knob, consistent with \citet{saunshi2025reasoning,bae2025relaxed}.\\
\newline
\textbf{Perception kernel.} GLA beats local softmax at single scale (129 vs.\ 140 PPL at 60k) but underperforms it in the hierarchical body (72.0 vs.\ 60.3); the right kernel depends on the schedule context. Full kernel ablation in Appendix~\ref{app:kernels}.

\subsection{Downstream Evaluation}
\label{sec:downstream}
\begin{table*}[t]
\centering
\small
\setlength{\tabcolsep}{6pt}
\resizebox{\textwidth}{!}{%
\begin{tabular}{lcccccccc}
\toprule
\textbf{Task}& \textbf{TF-6L@60k} & \textbf{TF-12L@60k} & \textbf{Hier-TextNCA} & \textbf{SWin-TF($w{=}128$)} & \textbf{Hier-Residual} & \textbf{Hier-NoStep} & \textbf{Hier-SinStep} & \textbf{UT(L5)} \\
 
\midrule
\multicolumn{9}{l}{\textit{Classification (\%, 3-seed mean):}} \\
\midrule
IMDb     & 90.84 & \textbf{91.00} & \textbf{91.00} & 90.26 & 89.82 & 90.07 & 87.92 & 89.25 \\
AG News  & 93.43 & \textbf{93.82} & 93.59 & 93.58 & 93.34 & 93.37 & 92.25 & 92.99 \\
SST-2    & 87.16 & 87.50 & \textbf{89.00} & 86.70 & 86.47 & 86.35 & 84.06 & 83.94 \\
\midrule
\textbf{Clf Avg} & 90.48 & 90.77 & \textbf{91.20} & 90.18 & 89.88 & 89.93 & 88.08 & 88.73 \\
\midrule
\multicolumn{9}{l}{\textit{Extractive QA (SQuAD v1.1):}} \\
\midrule
F1 & 28.6 & \textbf{33.2} & 26.3 & 22.4 & 23.6 & 22.8 & 19.4 & 27.4 \\
EM & 8.1  & \textbf{9.5}  & 7.1  & 5.8  & 6.3  & 5.9  & 4.9  & 7.4 \\
\midrule
\textbf{Params} & 35.5M & 54.5M & \textbf{30.8M} & 35.5M & 26.1M & 30.8M & 30.8M & 21.4M \\
\textbf{LM PPL} & 52.8  & 44.7  & \textbf{60.3}  & 63.2  & 76.6  & 82.7  & 119.5 & 91.7 \\
\bottomrule
\end{tabular}}
\caption{Hier-TextNCA is statistically tied with transformer baselines on classification, but underperforms on SQuAD. The QA gap tracks attention locality: global-attention models reach F1 27.4--33.2, while local-attention variants remain at F1 19.4--26.3.}
\label{tab:downstream}
\end{table*}
Table~\ref{tab:downstream} evaluates whether the language-modeling trends transfer to downstream tasks. On \textbf{classification}, Hier-TextNCA is competitive with transformer baselines but does not establish a clear transfer advantage: its 3-task average is 91.2\%, compared with 90.5\% for Transformer-6L and 90.8\% for Transformer-12L, and multi-seed results show overlapping error bars across AG News, SST-2, and IMDb. We treat classification as a statistical tie rather than a downstream win.

\textbf{Extractive QA} reveals a sharper limitation. Hier-TextNCA achieves an F1 score of 26.3 on SQuAD, below Transformer-6L (28.6) and Transformer-12L (33.2). More importantly, the failure mode aligns with attention locality: local-attention variants cluster at F1 19--26, whereas global-attention models cluster at F1 27--33. This suggests that the QA bottleneck lies in local context access, not in iterative updating itself. A hybrid readout variant with one added global-attention block also fails to close the gap, reaching 22.4 F1 despite increasing the model to 34.0M parameters (Appendix~\ref{app:downstream_full}, Figure~\ref{fig:downstream}). Thus, TextNCA transfers competitively to classification, but span-level QA remains limited by its local perception design. 

\section{Interpretability Analysis}
\label{sec:interp}

The staged design exposes where computation occurs. We summarise the main findings here; per-step language-modeling loss decomposition, within-stage attention maps, gate dynamics, and logit-lens analyses are reported in Appendix~\ref{app:interp_extra}.

\begin{figure}[hbt!]
\centering
\includegraphics[width=\columnwidth]{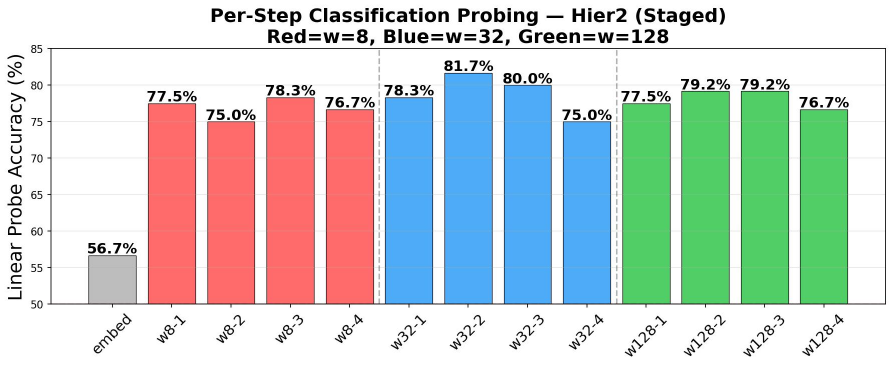}
\caption{Per-step linear probing accuracy across the 12 NCA steps (embedding shown as step 0). A 5-way topic classifier trained on hidden states at each step shows that representation quality is non-monotone: accuracy jumps from $56.7\%$ at the embedding to $77.5\%$ after the $w{=}8$ stage, peaks at $81.7\%$ during $w{=}32$, then drops to $79.2\%$ through $w{=}128$.}
\label{fig:probing}
\end{figure}

\textbf{Representation quality is non-monotone.} A 5-way topic linear probe on hidden states at each NCA step (Figure~\ref{fig:probing}, held-out WikiText-103 split) shows a non-monotone profile across the schedule. The middle stage is the most informative for sentence-level topic structure; the final wide-context stage trades some of that linear separability for next-token decode readiness.

\textbf{Loss reduction is concentrated in stage 3.} Decoding each hidden state via the shared LM head (Appendix~\ref{app:loss_per_step}, Figure~\ref{fig:loss_per_step}; 64 WikiText-103 sequences, 32{,}768 tokens) shows that next-token loss falls sharply only in the final stage: $\mathrm{S}3.2\!\to\!\mathrm{S}3.3\!\to\!\mathrm{S}3.4$ reduce loss by $2.4+1.9+1.5=5.8$ nats, while stages 1--2 plateau with per-step changes between $-0.3$ and $+0.2$ nats. The middle stage, despite being the probing peak, contributes no measurable LM-loss reduction: probing accuracy and next-token loss measure different aspects of the computation.

\begin{figure}[t!]
\centering
\includegraphics[width=\columnwidth]{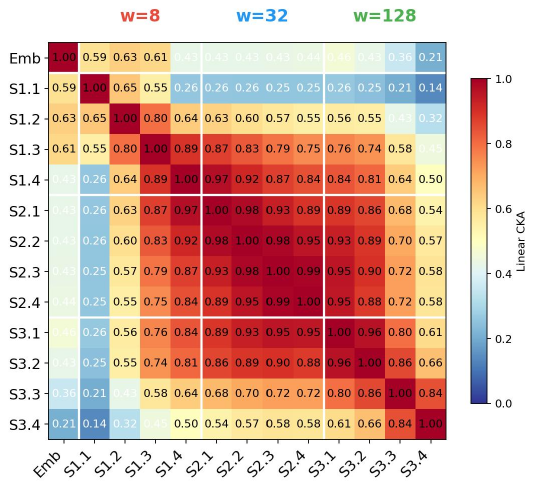}
\caption{Linear CKA between the embedding state and all 12 NCA steps (2{,}048 WikiText-103 tokens; random-pair floor $\approx 0.20$). White lines mark stage transitions. S1.1 and S3.4 are the only strong outliers; states S1.4--S3.1 form a continuous cluster with pairwise CKA $>0.85$.}
\label{fig:cka_full}
\end{figure}

\textbf{Three regimes, visible in representation space.} Linear CKA between the embedding and the 12 NCA states (Figure~\ref{fig:cka_full}; \citealp{kornblith2019similarity}) shows block structure driven by two outliers, S1.1 and S3.4, with the intermediate band S1.4--S3.1 forming a single tight cluster. The qualitative transitions happen at the schedule's endpoints, not at every stage boundary; this corroborates the probing finding.

\textbf{Receptive-field $\times$ stage mechanism.} Within-stage attention maps (Appendix~\ref{app:attn_progression}, Figure~\ref{fig:attn_progression}) give a direct mechanism: attention from ``Germany'' to ``France'' (a 7-token cross-clause association) is $\leq 9\%$ throughout stages 1--2, where the argmax is the adjacent token ``capital'', and jumps to $20$--$27\%$ in stage 3 with ``France'' becoming the argmax. The cross-clause completion requires a window that reaches the 7-token distance, which first happens at $w{=}128$.

\textbf{Prepare, then decode.} Together these views support a \emph{prepare-then-decode} reading: stages 1--2 build a linearly separable representation without committing to a specific next token; stage 3 then performs the cross-clause associations and converts the prepared representation into next-token probability mass. The contribution of stages 1--2 is representation preparation rather than direct prediction. Appendix~\ref{app:interp_extra} reports remaining evidence (GRU gate accumulate--consolidate--rewrite pattern, logit-lens trajectories) consistent with this reading.
\section{Test-time iteration-count control}
\label{sec:ttc}

The iterated weight-shared design admits a unique knob: varying inference iteration count $T_s'$ without retraining. The deterministic flagship has no usable knob: PPL is U-shaped in $T_s'$, bottoming at the training value and diverging (PPL collapses to a degenerate regime, $\gg 10^3$, where exact values are uninformative) for $T_s' \in \{1, 8, 10, 12\}$, because learned per-step embeddings make off-distribution $T_s'$ values out-of-distribution conditioning signals and $\|\Delta\mathbf{h}\| \approx 4$ even at step 12 (no fixed point).

Training a variant with sinusoidal step embeddings and $T_s \sim \mathrm{Uniform}\{2,4,6\}$ restores a working knob: \textsc{StochasticT} maintains PPL 101--108 across $T_s'\in\{2,4,6\}$ and partially generalises to $T_s'{=}5$ (PPL 164), while the deterministic SinStep counterpart diverges at any off-distribution $T_s'$ (PPL $\gg 10^3$). The cost is substantial: StochasticT's best PPL (101) is 41 worse than the deterministic flagship (60.3), partly because sinusoidal step embeddings underperform learned ones by $\sim 59$ PPL even at fixed $T_s$ (Table in Appendix~\ref{app:ttc_full}), and partly because averaging over a distribution of effective depths appears to reduce specialisation. The result demonstrates that an inference-time iteration-count knob is achievable at $\sim 30$M parameters with iterated weight-shared LMs; we deliberately avoid the term ``test-time compute scaling,'' which usually denotes inference-time search or chain-of-thought procedures and would overclaim what we show.

\section{Conclusion}
\label{sec:conclusion}

We started from a simple question: can an NCA-style computation, with local perception, weight-shared iteration, and a gated update, work as a language model, and which of those ingredients does the real work? The model we build runs, but most of its behaviour is explained by a single design choice, the staged window schedule that begins narrow and widens. Iteration and weight sharing add a smaller benefit on top, and only when the gate and the learned per-step embeddings are kept in place. Looking inside the model gives a consistent picture. The first stages do not directly lower next-token loss; they reorganise the hidden state, and the loss falls when the final wide stage finally decodes that representation. Iteration behaves like a depth knob with a clear optimum, and adding more steps eventually hurts performance. Training with random iteration counts gives a working inference-time depth knob, at the cost of higher absolute perplexity. Transformers still come out ahead on language modelling at this scale, and the gap widens once compute is matched, so the contribution here is analytical: a controlled reading of which parts of NCA-style computation actually carry the weight.

\section{Limitations}
\label{sec:limitations}
Whether iteration becomes a first-order contributor at substantially larger scales \citep{geiping2025latent} is open; our setting at $\sim 30$M parameters does not constrain that question and we do not report a parameter--PPL frontier curve. The matched-step comparison is the most favourable framing; under matched compute the gap to Transformer-6L widens substantially (Appendix~\ref{app:flops_matched}, Table~\ref{tab:flops_matched}). TextNCA is not computationally Pareto-superior at this scale; any ``parameter efficiency'' reading is restricted to matched-parameter, matched-step protocols.

Training on SlimPajama with Mistral 32K BPE and evaluating on WikiText-103 is an out-of-domain protocol; our perplexities are not comparable to literature numbers under the standard in-domain plus adaptive-softmax setting (see \S\ref{sec:lm_eval}), and all comparisons here are restricted to baselines trained under the same SlimPajama--Mistral protocol. Three-seed evaluation confirms the apparent classification advantage over Transformer-6L is within seed-level noise; no robust transfer claim should be inferred. The SQuAD gap tracks attention locality rather than iteration, so closing it requires global attention within the stack.

The inference-time iteration-count knob requires a substantial absolute-PPL penalty (101 vs.\ 60.3) and rules out an iterative-refinement reading of $T_s$. We do not include ALBERT or Mamba baselines under our protocol.

\section{Ethical Consideration}
\label{sec:ethics}

This work is a small-scale analytical study of a language-model architecture and does not introduce new data, deployed systems, or capabilities beyond those standard at the parameter scale we study. All experiments use public research benchmarks (SlimPajama, WikiText-103, IMDb, AG News, SST-2, SQuAD~v1.1) for their established research purposes. Total project compute is approximately 80 GPU-hours on two NVIDIA RTX PRO 6000 Blackwell GPUs; per-run costs are itemised in Appendix~\ref{app:compute}, Table~\ref{tab:compute}. We release training code, configurations and checkpoints under an open-source license (Appendix~\ref{app:reproducibility}); released models are small ($\leq 55$M parameters) and intended for research use only. We do not foresee dual-use or safety concerns beyond those common to academic language-modelling research at this scale.

\bibliography{references}  
%
%

\appendix

\section{Full Architecture Specification}
\label{app:arch_full}

This appendix provides the detailed update equations, perception-kernel and gate variants, and non-hierarchical orchestration strategies referenced in \S\ref{sec:architecture}.

\subsection{The NCA primitive in context}
\label{app:nca_primitive}

A Neural Cellular Automaton is a grid of cells, each holding a state vector and updated synchronously by a shared local rule. In the image variant of \citet{mordvintsev2020growing}, each cell sees a $3{\times}3$ neighbourhood through a depthwise convolution, passes the perception output through a small MLP, and writes the result back through a stochastic ``fire rate'' that gates the residual update. Three properties are load-bearing. Weight sharing: all cells use the same parameters at every step. Locality: each cell only sees its immediate neighbours. Iteration: the same rule is applied $T$ times so that global structure emerges from repeated local interactions \citep{mordvintsev2020growing,wolfram2002new}.

TextNCA preserves these three properties for autoregressive language modelling. Each token position $i$ holds a state $\mathbf{h}_i \in \mathbb{R}^d$; the 2D convolution is replaced by 1D causal windowed attention with window $w$; the fire rate is replaced by a GRU update gate (\S\ref{app:gates}). The per-iteration update is in-place and shared across iterations within a stage. The mapping to image-domain primitives is summarised in Table~\ref{tab:nca-mapping}. The hierarchical variant introduces three stages with their own parameter sets, so the rule is piecewise constant across stages while still being a shared local rule within each stage; this is what we mean when we describe the model as a hierarchical NCA rather than a single monolithic CA.

\subsection{Core update equations}
\label{app:core_update}

Given input tokens $\mathbf{x} = (x_1, \ldots, x_L)$, initial states are
{\small
\begin{equation}
\mathbf{h}_0 = \text{TokenEmbed}(\mathbf{x}) + \text{PosEmbed}(\mathbf{x}) \in \mathbb{R}^{B \times L \times d}.
\end{equation}
}
At each iteration $t \in \{1, \ldots, T\}$,
{\small
\begin{equation}
\mathbf{h}_t = \text{Gate}\!\left(\mathbf{h}_{t-1},\; \text{FFN}\!\left(\text{LN}\!\left(\text{Perceive}(\mathbf{h}_{t-1} + \mathbf{s}_t)\right)\right)\right),
\label{eq:app_step}
\end{equation}
}
where $\mathbf{s}_t \in \mathbb{R}^d$ is a learnable step embedding and all parameters (Perceive, FFN, Gate, LN) are shared across all $T$ iterations. Final predictions are
\begin{equation}
\text{logits} = \text{LMHead}(\text{LN}(\mathbf{h}_T)).
\end{equation}

In the hierarchical staged variant the per-iteration sharing holds \emph{within each stage}; the rule changes at stage boundaries. For each stage $k = 1..K$, the hidden state is refined $T_s$ times:
{\small
\begin{align}
\text{For } t &= 1..T_s: \nonumber \\
\mathbf{u} &= \text{FFN}_{\theta_k}(\text{LN}_{\theta_k}(\text{Perceive}_{\theta_k, w_k}(\mathbf{h} + \mathbf{s}_{(k-1)T_s + t}))) \\
\mathbf{h} &\leftarrow \text{Gate}_{\theta_k}(\mathbf{h}, \mathbf{u}).
\end{align}
}
The output of stage $k$ becomes the input to stage $k{+}1$ with no reset.

\subsection{Perception kernels}
\label{app:kernels}

All main experiments use local softmax attention; three other kernels are used in the perception-kernel ablation (\S\ref{sec:component_ablations}).

\paragraph{Local softmax attention} (main). Standard multi-head attention restricted to a causal window of size $w$:
{\small
\begin{equation}
\text{Attn}(Q_i, K, V) = \text{softmax}\!\left(\frac{Q_i K_{[i-w+1:i]}^\top}{\sqrt{d_h}}\right) V_{[i-w+1:i]}.
\end{equation}
}

\paragraph{Local linear attention.} Approximates softmax via a Taylor-2 feature map $\phi(\mathbf{x}) = [1, \mathbf{x}, \mathbf{x} \otimes \mathbf{x}/\sqrt{2}]$, applied within the local window.

\paragraph{Gated linear attention (GLA).} Following \citet{yang2024gated}, introduces data-dependent decay gates per head:
{\small
\begin{equation}
\mathbf{g}_j^{(h)} = \sigma(\mathbf{W}_g^{(h)} \mathbf{K}_j), \quad \text{decay}_{j \to i} = \prod_{s=j+1}^{i} (1 - \mathbf{g}_s^{(h)}).
\end{equation}
}

\paragraph{Convolution.} 1D causal depthwise convolution with kernel size $w$.

\paragraph{Single-scale results.}
At single scale ($w{=}32$, $T{=}12$) the kernel ordering on WikiText-103 PPL at 60k steps is GLA (129) $<$ local softmax (140) $<$ convolution (172) $<$ local linear (176). Within the hierarchical body the ordering inverts: GLA reaches 72.0 PPL versus local softmax's 60.3. The interaction is consistent with GLA's data-dependent decay gates providing the largest benefit at small receptive fields where softmax attends uniformly; at $w \in \{32, 128\}$ softmax already achieves sharp query-dependent attention, and GLA's scalar-decay bottleneck becomes a limitation.

\begin{figure}[t]
\centering
\includegraphics[width=\columnwidth]{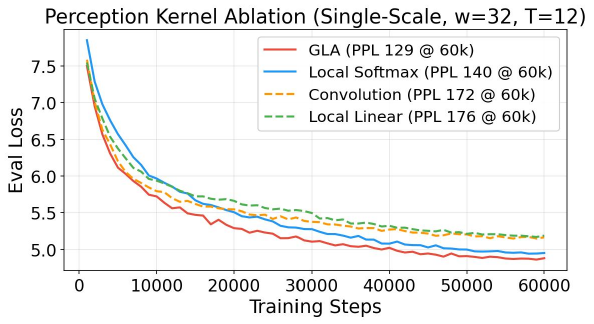}
\caption{Perception-kernel ablation at single scale ($w{=}32$, $T{=}12$), all kernels clipped to 60k training steps. GLA (129 PPL) clearly beats local softmax (140), while convolution (172) and local linear (176) lag by 32--36 PPL. The hierarchical body inverts this ranking, with local softmax reaching 60.3 PPL and GLA 72.0.}
\label{fig:kernel_ablation}
\end{figure}

\subsection{Gating mechanisms}
\label{app:gates}

All main experiments use a GRU gate; four ablation comparators are reported in \S\ref{sec:component_ablations}.

\paragraph{GRU gate} (main). The gate controls how the perception output $\mathbf{u}$ combines with the previous state $\mathbf{h}_{t-1}$:
\begin{align}
\mathbf{z}      &= \sigma(\mathbf{W}_z[\mathbf{h}_{t-1}; \mathbf{u}]) & \text{(update)} \\
\mathbf{r}      &= \sigma(\mathbf{W}_r[\mathbf{h}_{t-1}; \mathbf{u}]) & \text{(reset)} \\
\tilde{\mathbf{h}} &= \tanh(\mathbf{W}_h[\mathbf{r} \odot \mathbf{h}_{t-1}; \mathbf{u}]) & \text{(candidate)} \\
\mathbf{h}_t    &= (1 - \mathbf{z}) \odot \mathbf{h}_{t-1} + \mathbf{z} \odot \tilde{\mathbf{h}} & \text{(output)}.
\end{align}
Ablation comparators: \textbf{sigmoid} ($g \odot u + (1{-}g) \odot h$), \textbf{highway} (gate depends only on $h$), \textbf{residual} ($h + u$), \textbf{overwrite} ($u$).

\paragraph{Gating-mechanism ablation.}
We replace the GRU gate with simpler combiners while keeping all other architecture and training settings fixed; Table~\ref{tab:gating_full} reports the result. The GRU gate is the only combiner that beats the residual baseline by a clear margin on PPL.
\begin{table}[h]
\centering
\small
\setlength{\tabcolsep}{4pt}
\begin{tabular}{lrrrr}
\toprule
\textbf{Gate} & \textbf{Params} & \textbf{PPL} & \textbf{Clf Avg} & \textbf{QA F1} \\
\midrule
GRU             & 30.8M & \textbf{60.3} & \textbf{91.2} & \textbf{26.3} \\
Residual ($+$)  & 26.1M & 76.6 & 89.9 & 23.6 \\
SigmoidDropout  & 27.7M & 71.7 & 89.8 & 23.4 \\
\bottomrule
\end{tabular}
\caption{Gating ablation at staged $[8,32,128]{\times}T_s{=}4$. The plain sigmoid combiner ($g \odot u + (1{-}g) \odot h$) diverges at step 3k due to unbounded state growth and is therefore omitted from the table.}
\label{tab:gating_full}
\end{table}

\paragraph{Step-embedding ablation.}
We replace the learned per-iteration embedding $\mathbf{s}_t$ with either no step embedding or a sinusoidal one, keeping all other architecture and training settings fixed; Table~\ref{tab:step_embedding_full} reports the result. Some form of parametric per-step conditioning is necessary for the iterated body to work, and the sinusoidal choice in particular is a poor fit under deterministic short-horizon $T_s$.
\begin{table}[h]
\centering
\small
\setlength{\tabcolsep}{4pt}
\begin{tabular}{lrrr}
\toprule
\textbf{Step embedding} & \textbf{PPL} & \textbf{Clf Avg} & \textbf{QA F1} \\
\midrule
Learned (flagship) & \textbf{60.3} & \textbf{91.2} & \textbf{26.3} \\
None              & 82.7$^{*}$ & 89.9 & 22.8 \\
Sinusoidal        & 119.5 & 88.1 & 19.4 \\
\bottomrule
\end{tabular}
\caption{Step-embedding ablation. $^{*}$Training interrupted; best checkpoint at step 48k.}
\label{tab:step_embedding_full}
\end{table}

\paragraph{$T_s$ sweep.}
We retrain Hier-TextNCA from scratch at $T_s \in \{2,4,6,8\}$ with all other architecture and training settings fixed; Table~\ref{tab:ts_full} reports the result. PPL is sharply U-shaped with a minimum at $T_s{=}4$, ruling out a monotone ``more iterations helps'' reading at fixed parameters.
\begin{table}[h]
\centering
\small
\setlength{\tabcolsep}{4pt}
\begin{tabular}{cccr}
\toprule
$T_s$ & Total NCA steps & Params & PPL \\
\midrule
2 & 6  & 30.83M & 80.3 \\
\textbf{4} & \textbf{12} & 30.84M & \textbf{60.3} \\
6 & 18 & 30.86M & 73.9 \\
8 & 24 & 30.87M & 117.9 \\
\bottomrule
\end{tabular}
\caption{Effect of iterations per stage $T_s$ at matched compute. PPL is U-shaped with a sharp minimum at $T_s{=}4$.}
\label{tab:ts_full}
\end{table}

\subsection{Non-hierarchical orchestration variants}
\label{app:non_hier}

For completeness, three non-hierarchical orchestrations used only as ablation comparators (Table~\ref{tab:orchestration} in the main paper):

\paragraph{Single-Layer TextNCA.} One perception kernel at window $w$, applied $T$ times.

\paragraph{MultiLayer TextNCA.} $N$ distinct sub-layers per iteration (e.g.\ 3 sub-layers $\times$ $T{=}4$), same window size, different projection weights per sub-layer.

\paragraph{Stacked NCA.} $N$ distinct cells (different weights) chained, each iterating $T_{\text{local}}$ times (variants 6C$\times$T2, 12C$\times$T1, etc.). All three score substantially worse than the hierarchical staged variant; see Table~\ref{tab:orchestration}.

\paragraph{Model configuration summary.}
\begin{table*}[h]
\centering
\small
\setlength{\tabcolsep}{4pt}
\begin{tabular}{lrrl}
\toprule
\textbf{Model} & \textbf{Params} & \textbf{Steps} & \textbf{Key design} \\
\midrule
TextNCA-L1        & 21.4M & 60k & $w{=}32$, $T{=}12$ \\
TextNCA-GLA       & 21.4M & 60k & GLA kernel \\
TextNCA-Conv      & 29.0M & 60k & Conv kernel \\
Hier-TextNCA      & 30.8M & 60k & Staged $[8,32,128]{\times}T_s{=}4$ \\
SWin-TF-Staged    & 35.5M & 60k & Layer windows $[8,8,32,32,128,128]$ \\
SWin-TF-6L        & 35.5M & 60k & Uniform $w{=}128$, 6 layers \\
Transformer-6L    & 35.5M & 60k & 6 distinct layers, global attn \\
Transformer-12L   & 54.5M & 60k & 12 distinct layers, global attn \\
UT                & 21.4M & 60k & Shared weights, global attn \\
\bottomrule
\end{tabular}
\caption{Model configurations. All TextNCA variants use $d{=}512$, $d_\text{ff}{=}2048$, $H{=}8$ heads, $d_h{=}64$.}
\label{tab:configs_full}
\end{table*}

\section{Full Downstream Results}
\label{app:downstream_full}

\begin{figure}[t]
\centering
\includegraphics[width=\columnwidth]{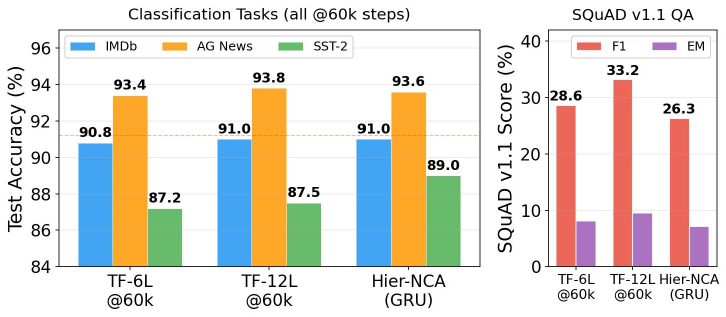}
\caption{Downstream evaluation at matched 60k training steps. Left: classification accuracy across IMDb, AG News and SST-2; Hier-TextNCA (30.8M) is within noise of both transformer baselines, the dashed line marks its 91.2\% three-task average. Right: SQuAD v1.1 F1; the gap separates global-attention models (F1 27.4--33.2) from local-attention variants (F1 19.4--26.3).}
\label{fig:downstream}
\end{figure}

\paragraph{Multi-seed classification.}
On AG News, Hier-TextNCA reaches $93.46 \pm 0.17$ versus Transformer-6L's $93.47 \pm 0.24$ (tied). On SST-2, Hier-TextNCA reaches $87.61 \pm 1.46$ versus Transformer-6L's $87.10 \pm 0.57$ (overlapping error bars). On IMDb, Hier-TextNCA reaches $90.76 \pm 0.18$ versus Transformer-6L's $90.40 \pm 0.25$ (within noise). We report this as a statistical tie rather than as a transfer advantage.

\paragraph{Hier-Hybrid: a global-attention readout block does not close the QA gap.}
We tested whether appending a single causal global-attention block plus FFN to the Hier-TextNCA stack at the readout would close the QA gap while preserving the iterated weight-shared local-attention body. Hier-Hybrid (34.0M parameters, $+3.2$M from the global block) reaches LM PPL 61.0 (essentially unchanged from Hier-TextNCA's 60.3), classification average 89.9 ($-1.3$ from 91.2), and SQuAD F1 22.4 (\emph{worse} than Hier-TextNCA's 26.3). The local-attention body's hidden states are not in a representation that the appended global block can productively rewrite; closing the QA gap requires a global-attention component \emph{within} the stack, not appended on top.

\section{Training dynamics}
\label{app:training_dynamics}

\begin{figure*}[t]
\centering
\includegraphics[width=\textwidth]{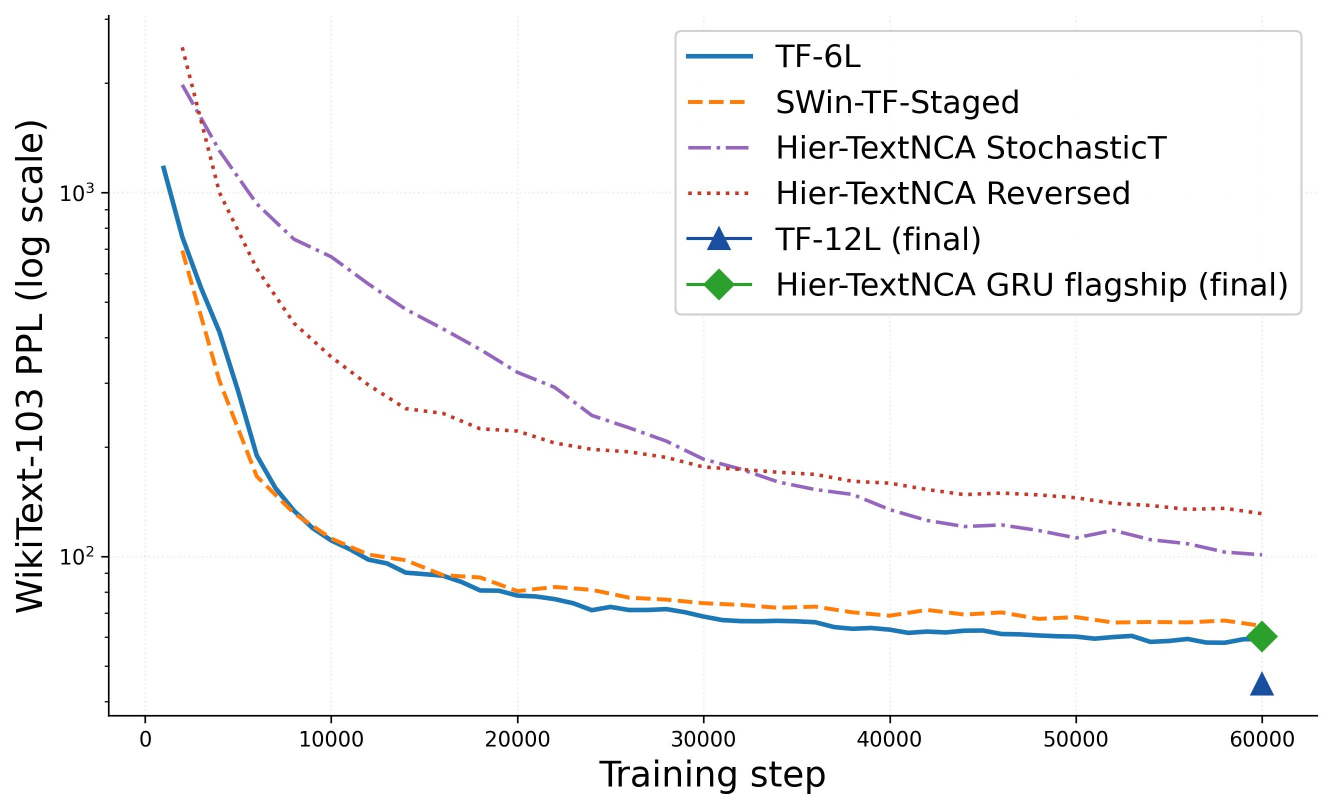}
\caption{WikiText-103 PPL (log scale) against training steps for the key TextNCA and Transformer variants. Hier-TextNCA GRU flagship (green diamond, PPL 60.3) and TF-12L (blue triangle, PPL 44.7) are final-step-only points; their continuous logs were not retained. TF-6L (blue) uses the full 120k-step log clipped to 60k. SWin-TF-Staged (orange, 64.5) tracks Hier-TextNCA closely, while Hier-TextNCA StochasticT (purple, 101) and Hier-TextNCA Reversed (red, 131) illustrate the cost of stochastic-$T_s$ training and of reversing the window schedule.}
\label{fig:eval_ppl}
\end{figure*}

Figure~\ref{fig:eval_ppl} plots WikiText-103 PPL against training steps for the key variants in Table~\ref{tab:main}, referenced from \S\ref{sec:main_lm}. Hier-TextNCA GRU and TF-12L are shown as final-step points because their continuous logs were not retained; the remaining curves are full training trajectories clipped to the 60k matched-step budget. SWin-TF-Staged tracks Hier-TextNCA closely throughout training and settles within $4.1$ PPL of it at 60k, consistent with the schedule reading in \S\ref{sec:schedule}. The Hier-TextNCA Reversed run is also a full curve: its PPL falls early and then plateaus much higher (131), so reversing the window order is a different solution basin and not a slower start that catches up. Hier-TextNCA StochasticT trains stably at a higher PPL throughout, the price of the iteration-count knob discussed in \S\ref{sec:ttc}.

\section{Additional Interpretability Analyses}
\label{app:interp_extra}

This appendix expands the main-paper interpretability analysis in \S\ref{sec:interp}. All analyses use the same Hier-TextNCA checkpoint; the running prompt for per-token figures is ``Paris is the capital of France and Berlin is the capital of Germany today.'', chosen because the demonstration pair Paris--France lies $7$ tokens before the next-token position, inside $w{=}128$ but outside $w{=}32$. The goal is not to assign fixed linguistic labels to stages, but to test whether the staged computation produces distinct internal regimes and whether those regimes explain the per-step loss decomposition discussed in \S\ref{sec:interp}.

\subsection{Per-step language-modeling loss}
\label{app:loss_per_step}

\begin{figure*}[hbt!]
\centering
\includegraphics[width=0.70\textwidth]{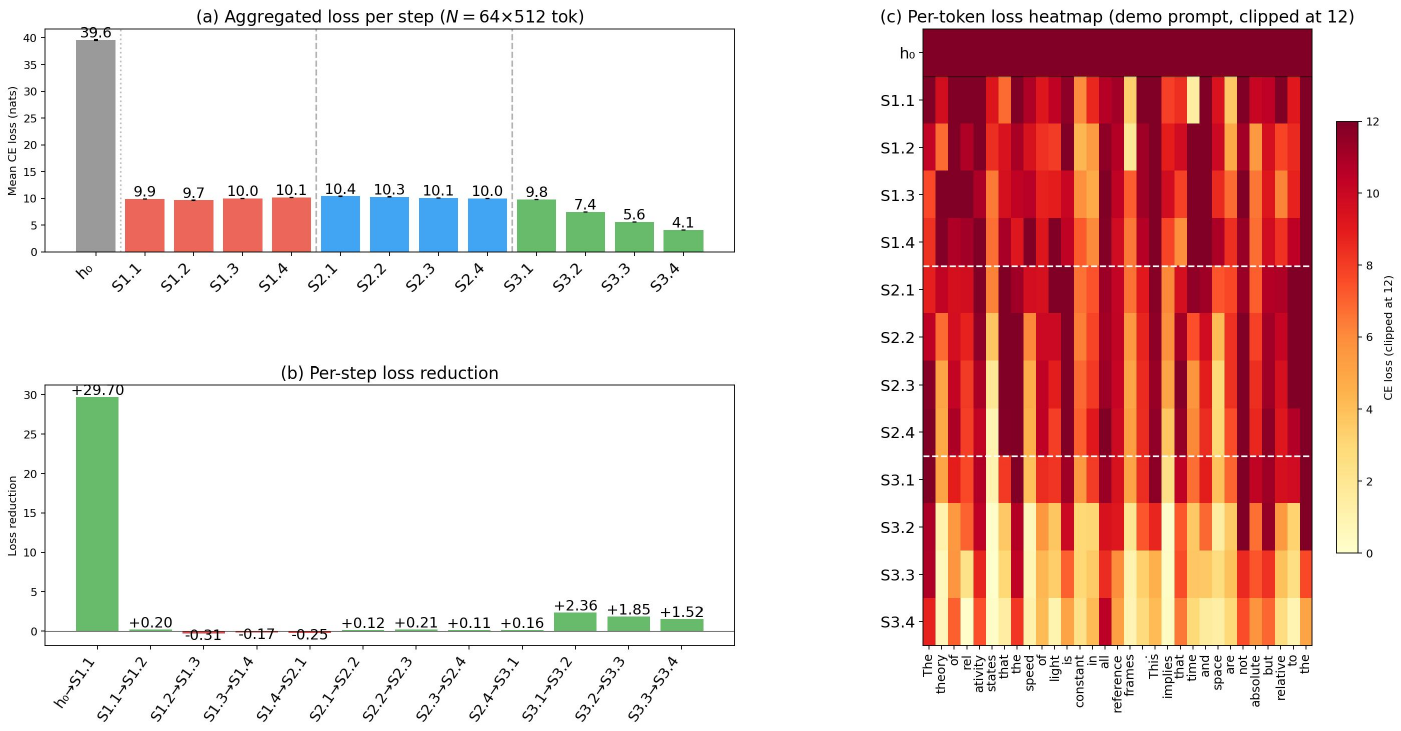}
\caption{Per-step next-token loss of Hier-TextNCA. \textbf{(a)} Mean cross-entropy over 32{,}768 WikiText-103 tokens. The embedding loss is inflated by tied-embedding self-decoding; the first NCA step collapses loss near $\log V$; stages 1--2 plateau; the final stage-3 steps carry nearly all LM-loss reduction. \textbf{(b)} Per-step loss reduction. \textbf{(c)} Per-token loss heatmap, clipped at 12 nats.}
\label{fig:loss_per_step}
\end{figure*}

Figure~\ref{fig:loss_per_step} reports the per-step next-token cross-entropy referenced from the main paper. We analyse all 12 intermediate NCA states by applying the shared LM head after each step and measuring next-token cross-entropy on 64 WikiText-103 sequences (32{,}768 tokens).

The pattern is highly asymmetric. The embedding state $\mathbf{h}_0$ has trivially large loss (39.6 nats) because tied embeddings self-decode toward the current token rather than the next token. The first NCA step collapses this to $9.9$ nats, close to the uniform-vocabulary ceiling $\log V \approx 10.4$. Stages 1 and 2 then largely plateau, with per-step changes between $-0.3$ and $+0.2$ nats. Nearly all meaningful next-token improvement occurs in the final three steps of stage 3: $\mathrm{S}3.2\!\to\!\mathrm{S}3.3\!\to\!\mathrm{S}3.4$ reduce loss by $2.4+1.9+1.5=5.8$ nats, taking perplexity on this 32{,}768-token subset from $18{,}138$ to $59$ (within sampling variance of the $60.3$ reported on the full WikiText-103 eval set).

\subsection{Full representation-similarity matrix}
\label{app:cka}

Figure~\ref{fig:cka_full} (in the main paper) shows the complete $13{\times}13$ linear centred-kernel-alignment matrix over the embedding state and all 12 NCA states, estimated on 2{,}048 WikiText-103 tokens. The random-pair floor at this sample size is $\approx 0.20$; all values below should be read against that baseline rather than against zero.

Three concrete observations refine the picture. (i) The embedding row sits at $\approx 0.43$ to the bulk of the NCA states, so a single NCA step already moves representations out of the tied-embedding geometry. (ii) The first step S1.1 is a strong outlier: its similarity to every state after stage 1 is $\leq 0.26$, so the very first $w{=}8$ update is the largest qualitative rewrite in the pipeline. (iii) States S1.4 through S3.1 form a tight cluster with pairwise CKA $> 0.85$, so the $w{=}8 \to 32$ and $w{=}32 \to 128$ transitions are smooth continuations of the same regime rather than resets. The final state S3.4 is the second outlier: 0.84 to S3.3, but $\leq 0.58$ to any state earlier than S3.2, consistent with its decode-ready role identified by the gate analysis in \S\ref{app:gate_decomp}.

\subsection{Semantic clustering across stages}
\label{app:similarity}

\begin{figure*}[ht]
\centering
\includegraphics[width=\textwidth]{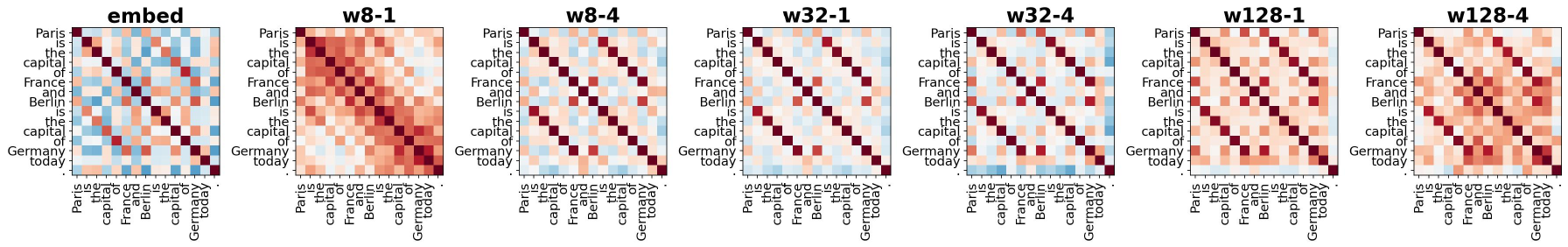}
\caption{Pairwise cosine similarity between the 15 token hidden states at seven snapshots (embedding + start and end of each of the three stages), colored on $[-0.5,1]$ (red high, blue low, diagonal $=1$). Paris--France $0.48\!\to\!0.73\!\to\!0.57$, Berlin--Germany $0.71\!\to\!0.85$, and the same-syntactic-role pair Paris--Berlin $0.54\!\to\!0.75$. The random-pair median also drifts up ($0.22\!\to\!0.43$); the two named pairs stay clearly above this baseline at every step.}
\label{fig:similarity}
\end{figure*}

Figure~\ref{fig:similarity} shows pairwise cosine similarity between the 15 token hidden states at seven snapshots (embedding plus the start and end of each stage), colored on $[-0.5, 1]$. Three observations on the running prompt. (i) Semantically related pairs rise across stages: Paris--France goes $0.48 \!\to\! 0.67$ (end of Stage 2) $\!\to\! 0.73$ (start of Stage 3) $\!\to\! 0.57$ (final step); Berlin--Germany starts already high at $0.71$ (Mistral's tied embeddings co-locate capital/country tokens, so most of the work is sharpening rather than discovery) and rises monotonically to $0.85$. (ii) Paris--Berlin, which share a syntactic role but not an entity, also climbs from $0.54$ to a peak of $0.75$ in Stage 3, indicating the staged computation aligns tokens by position-in-construction, not only by lexical identity. (iii) Global anisotropy grows with depth: the median off-diagonal similarity for arbitrary token pairs rises from $0.22$ (embedding) to $0.43$ (final step), so each pair should be compared against the same-step random baseline rather than against zero. Both named pairs stay well above this baseline at every step.

\subsection{GRU gate decomposition}
\label{app:gate_decomp}

\begin{figure}[ht]
\centering
\includegraphics[width=\columnwidth]{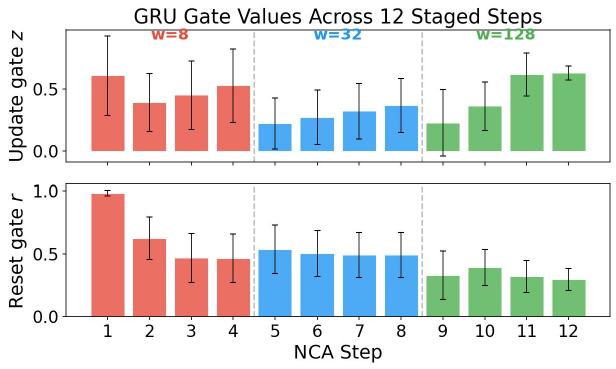}
\caption{GRU gate statistics per NCA step. Bars are means over hidden dimensions; error bars are $\pm 1\sigma$. Top: update gate $z$, a small U-shape in the $w{=}8$ stage ($0.60, 0.39, 0.52$), $\sim 0.30$ through the $w{=}32$ stage, and rising to $0.62$ by step 12. Bottom: reset gate $r$, falling monotonically from $0.98$ at step 1 to about $0.30$ in steps 9--12. The pair encodes an accumulate--consolidate--rewrite role across the three stages.}
\label{fig:gate_full}
\end{figure}

Figure~\ref{fig:gate_full} reports the update gate $z$ and reset gate $r$ averaged across hidden dimensions at every NCA step, with $\pm 1\sigma$ bands. The update gate has a small U-shape inside the $w{=}8$ stage (about $0.60$, $0.39$, $0.52$ at steps 1, 2 and 4), sits low and roughly flat at $\sim 0.30$ through the $w{=}32$ stage, and then rises sharply in the $w{=}128$ stage, reaching $0.62$ by step 12. The reset gate falls almost monotonically across the full 12-step trajectory: from $0.98$ at step 1, where the candidate is essentially computed from the existing state, down to about $0.30$ in steps 9--12, where the candidate is computed largely without reference to it.

Read together, the two gates give a coarse role for each stage. Stage 1 has $r$ near pass-through and $z$ moderate, so each step writes a modest update on top of the existing state; the model is \emph{accumulating} local context. Stage 2 has $z$ low and $r$ still falling, so the state is mostly carried forward with small adjustments; the model is \emph{consolidating}. Stage 3 has $r$ low and $z$ high, so the new candidate is computed with little regard for the previous state and is then written into it strongly; the model is \emph{rewriting} for next-token prediction. This matches the asymmetric loss curve in \S\ref{sec:interp}, where measurable LM-loss reduction is concentrated in the final three steps. We describe gate dynamics here, and not a fixed mapping of stages to linguistic categories.

\subsection{Within-stage attention progression}
\label{app:attn_progression}

\begin{figure*}[ht]
\centering
\includegraphics[width=\textwidth]{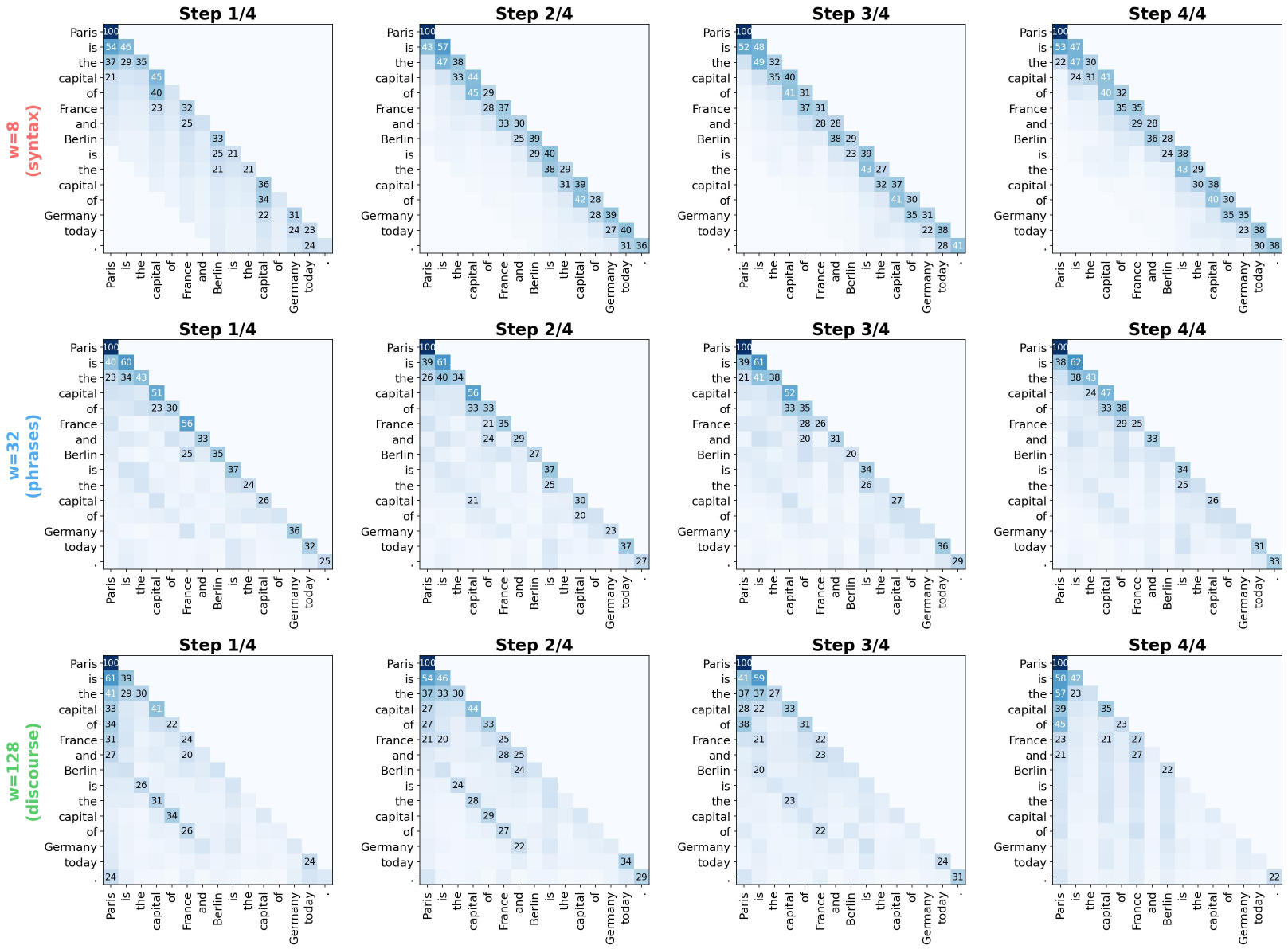}
\caption{Within-stage attention maps for the running prompt. Rows are stages ($w{=}8, 32, 128$); columns are iterations $1\!\to\!4$. Cell values are head-averaged attention weights shown as percentages (e.g.\ ``27'' means $0.27$); only cells with weight $>0.20$ are annotated. The cross-clause association ``Germany'' $\to$ ``France'' (7-token distance) is $\leq 9\%$ throughout Stages 1--2 and jumps to $20$--$27\%$ in Stage 3, becoming the argmax.}
\label{fig:attn_progression}
\end{figure*}

Figure~\ref{fig:attn_progression} shows head-averaged softmax attention maps within each stage on the running prompt. Each row is a stage (weights shared within a row, windows $w \in \{8, 32, 128\}$); each column is an iteration $1\!\to\!4$ within that stage. Despite identical weights, the attention pattern changes across iterations because the underlying hidden states evolve. The cross-clause association is the cleanest case: the attention from ``Germany'' to ``France'' is $\leq 0.09$ throughout Stages 1--2, where the argmax is the adjacent token ``capital''; it then jumps to $0.20$--$0.27$ in Stage 3, with ``France'' itself becoming the argmax. This is a direct attention-level instance of the receptive-field argument: the model can only perform the pattern-completion ``Berlin's capital-analogue'' once the window reaches the 7-token Berlin$\leftrightarrow$France distance, which first happens in the $w{=}128$ stage.

\subsection{Per-step linear probing}
\label{app:probing}

Figure~\ref{fig:probing} (in the main paper) plots the per-step linear-probe accuracy. A linear probe trained on hidden states at each NCA step (5-way topic classification on a held-out WikiText-103 split) reveals a clear non-monotone profile across stages: accuracy is $56.7\%$ at the embedding, jumps to $77.5\%$ after the $w{=}8$ stage, peaks at $81.7\%$ during the $w{=}32$ stage, and stabilizes at $79.2\%$ through $w{=}128$. Two readings follow. First, the mid-range $w{=}32$ stage is the most informative for sentence-level topic structure, while the final wide-context stage trades some of that linear separability for next-token decode readiness, consistent with the loss decomposition in \S\ref{app:loss_per_step}. Second, probing accuracy and LM loss therefore track different aspects of the computation: a stage can be highly probe-accurate (Stage 2) without contributing measurable next-token loss reduction.

\subsection{Hidden-state dynamics}
\label{app:hidden_dynamics}

\begin{figure}[ht]
\centering
\includegraphics[width=\columnwidth]{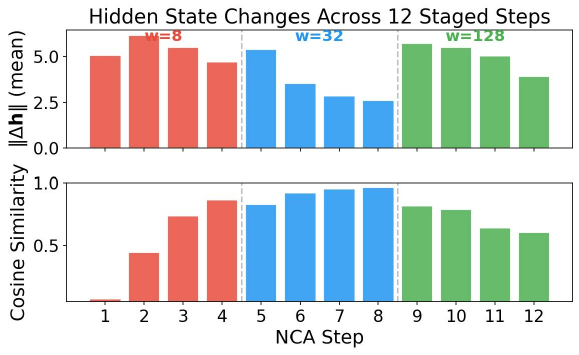}
\caption{Hidden-state dynamics across the 12 NCA steps. Top: update magnitude $\|\Delta\mathbf{h}\|$. Bottom: cosine similarity between consecutive states. States stabilize within stages but shift at stage boundaries; the final update remains nonzero.}
\label{fig:hidden_dynamics}
\end{figure}

Figure~\ref{fig:hidden_dynamics} tracks the update magnitude $\|\Delta \mathbf{h}\|$ and cosine similarity between consecutive states. Within each stage, cosine similarity increases, indicating local stabilization. At stage boundaries, similarity drops, indicating a shift to a new representational regime. Importantly, the update magnitude does not vanish by the final trained step, so the NCA is not simply converging to a fixed point within the trained horizon.

This helps interpret the iteration ablation: increasing the number of steps is not guaranteed to improve performance by further convergence. Instead, additional untrained iterations can move the hidden state outside the region that the LM head has learned to decode.

\subsection{Logit-lens evolution across stages}
\label{app:logit_lens}

\begin{figure*}[ht]
\centering
\includegraphics[width=\textwidth]{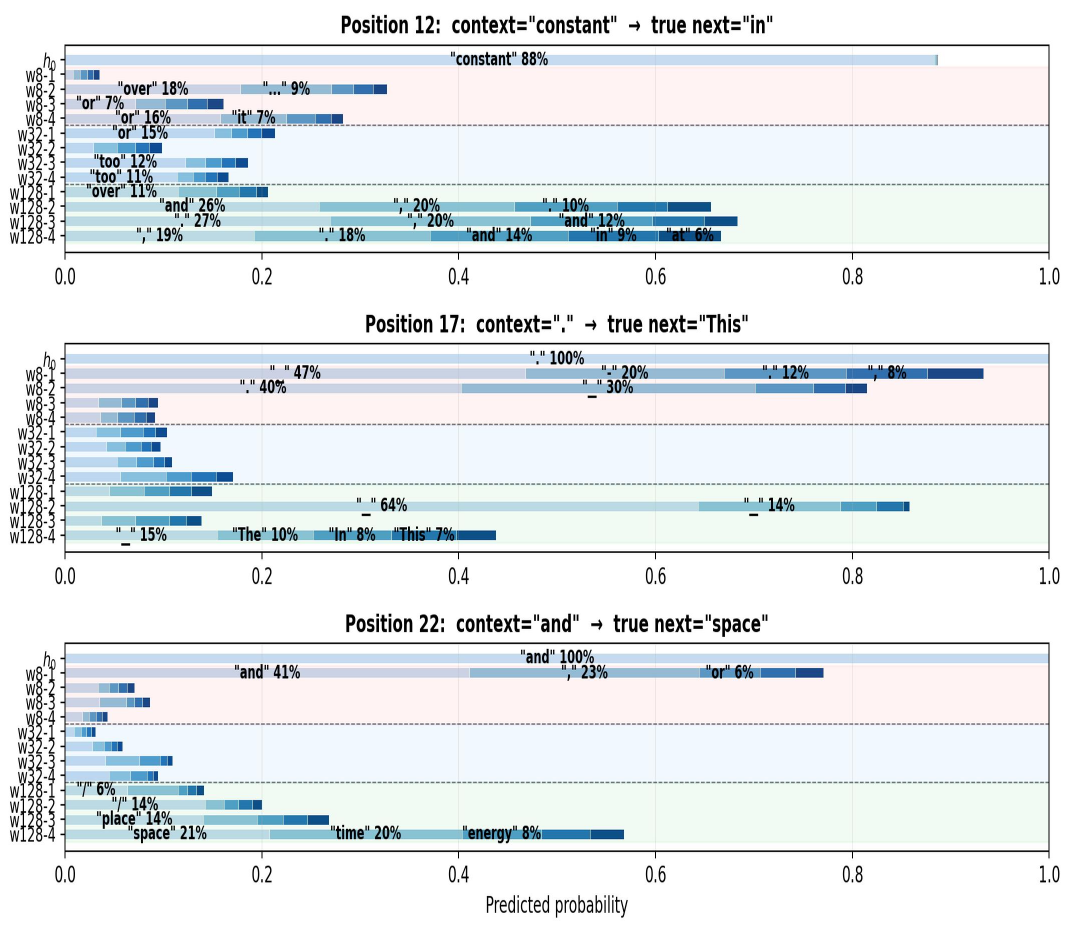}
\caption{Logit lens across the 13 representations (embedding + 12 NCA steps) for three positions in the relativity prompt. Each bar is the top-5 distribution at that step; shading marks stages (red $w{=}8$, blue $w{=}32$, green $w{=}128$). Position 12: ``constant''$\to$``in''. Position 17: ``.''$\to$``This''. Position 22: ``and''$\to$``space''. Stages 1--2 mostly rearrange representations; Stage 3 produces the usable next-token distribution.}
\label{fig:logit_lens}
\end{figure*}

Figure~\ref{fig:logit_lens} applies the LM head at each of the 13 representations (embedding $\mathbf{h}_0$ + 12 NCA steps) for three positions in the science prompt ``The theory of relativity states that the speed of light is constant in all reference frames. This implies that time and space are not absolute$\ldots$''. Each horizontal bar shows the top-5 predicted-token distribution at that step; background shading marks stages (red $w{=}8$, blue $w{=}32$, green $w{=}128$). Three positions illustrate the prepare-then-decode pattern. \textbf{Position 12} (``constant''$\to$``in''): $\mathbf{h}_0$ self-decodes as ``constant'' with 88\% probability; Stage 1 produces short local continuations (``over'', ``or'', ``it''); Stage 3 sharply shifts to function-word continuations with ``in'' reaching the top-5. \textbf{Position 17} (``.''$\to$``This''): after the sentence boundary, Stages 1--2 predict punctuation-like continuations (``.'', ``-''), while Stage 3 produces sentence-initial tokens ``The'' / ``In'' / ``This'' (7\%). \textbf{Position 22} (``and''$\to$``space''): only Stage 3 produces content words for the compound continuation, with ``space'' 21\%, ``time'' 20\%, ``energy'' 8\%. Across all three positions, Stages 1--2 rearrange representations and Stage 3 converts them into usable next-token distributions, consistent with the per-step loss analysis (Figure~\ref{fig:loss_per_step}).

\subsection{Per-token logit-lens trajectories}
\label{app:token_evolution}

\begin{figure}[ht]
\centering
\includegraphics[width=\columnwidth]{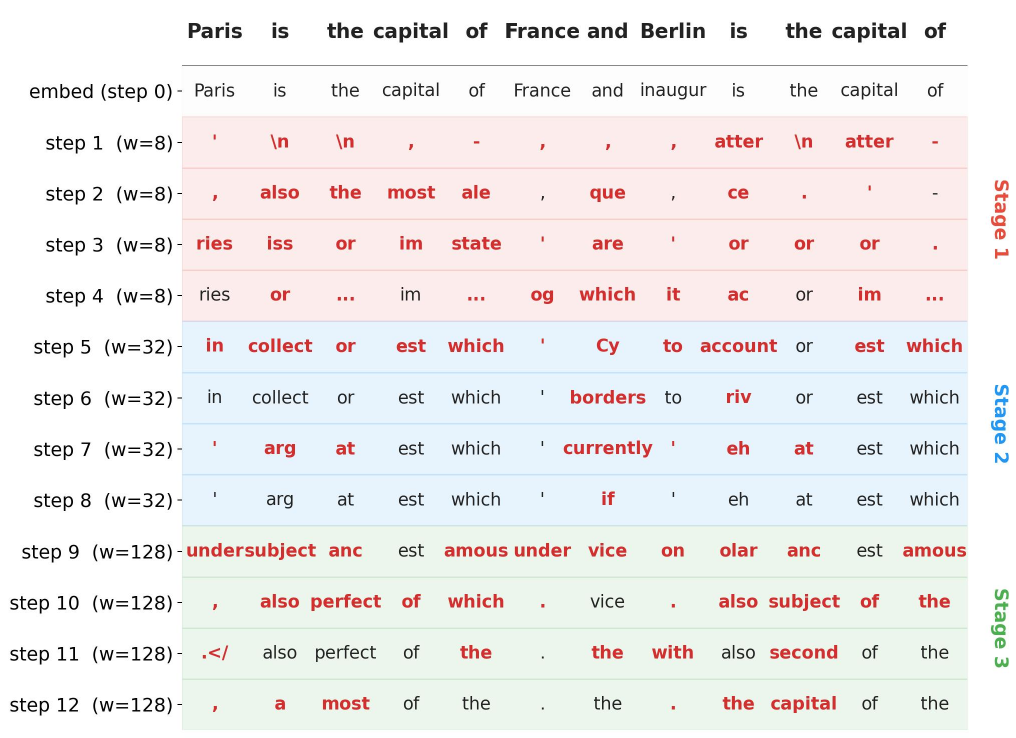}
\caption{Per-token logit-lens trajectories across 12 NCA iterations on the running prompt. Rows: embedding + 12 steps (stage-shaded). Columns: token positions. Each cell shows the top-1 decoded token from that position's hidden state; red marks a change from the previous step. Stage 1 produces filler, Stage 2 produces relational connectives, Stage 3 produces content words that match the true completion.}
\label{fig:token_evolution}
\end{figure}

Figure~\ref{fig:token_evolution} traces top-1 logit-lens decodings per token across the 12 NCA iterations on the running Paris--Berlin prompt. At step 0 each position decodes to itself (the embedding). In the $w{=}8$ stage, tokens drift to punctuation and filler subwords ($-$, $,$, \texttt{\textbackslash n}) because the 7-neighbour window is too narrow for factual retrieval. In $w{=}32$, positions acquire \emph{relational} roles (``which'', ``state'', ``being'', ``borders'') --- a syntactic scaffold shared across positions. Only in $w{=}128$ does content crystallize: the final position settles on ``the'' (56\%) with ``France'', ``Europe'', ``Paris'', and ``Germany'' in ranks 2--5. The 7-token distance Paris$\leftrightarrow$France is inside $w{=}128$ but outside $w{=}32$, so the factual pattern completion is only possible once the wide-context stage is reached. This figure should be read as a mechanistic illustration, not a benchmark; a stronger factual-association claim would require a probe across many prompts.

\subsection{Per-position prediction confidence}
\label{app:pred_evolution}

\begin{figure}[ht]
\centering
\includegraphics[width=\columnwidth]{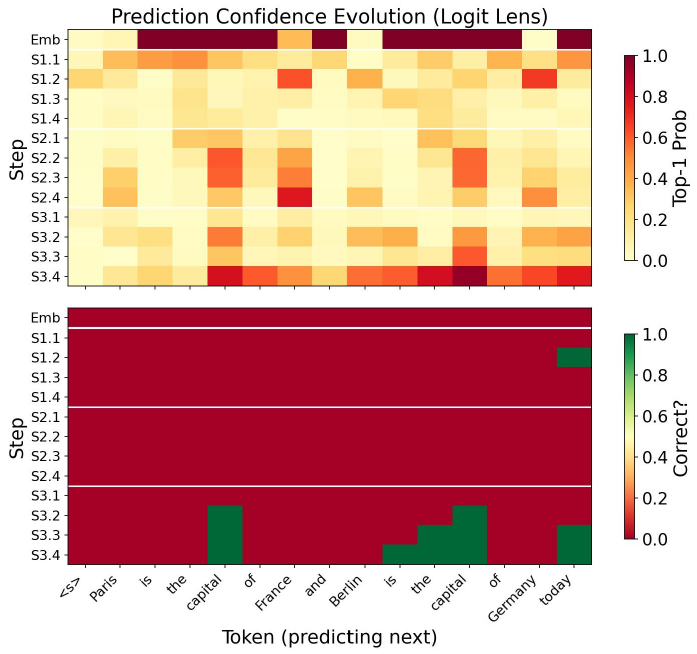}
\caption{Per-position logit-lens prediction evolution. Top: top-1 probability. Bottom: whether the top-1 prediction matches the true next token. Correct predictions emerge primarily in stage 3.}
\label{fig:pred_evolution}
\end{figure}

Figure~\ref{fig:pred_evolution} gives a per-token view of the prepare-then-decode pattern. Top-1 confidence is high at the embedding state because the tied LM head can decode the input token, but this confidence is not useful for next-token prediction. Confidence then remains low through most of stages 1 and 2, before correct next-token predictions emerge in stage 3 for positions whose continuation is locally well determined.

This complements the aggregate loss curve: early stages destroy the trivial copy prediction and reshape the representation, while stage 3 turns the representation into next-token probability mass.

\subsection{Top-$k$ evolution at the final position}
\label{app:topk_final}

\begin{figure}[ht]
\centering
\includegraphics[width=\columnwidth]{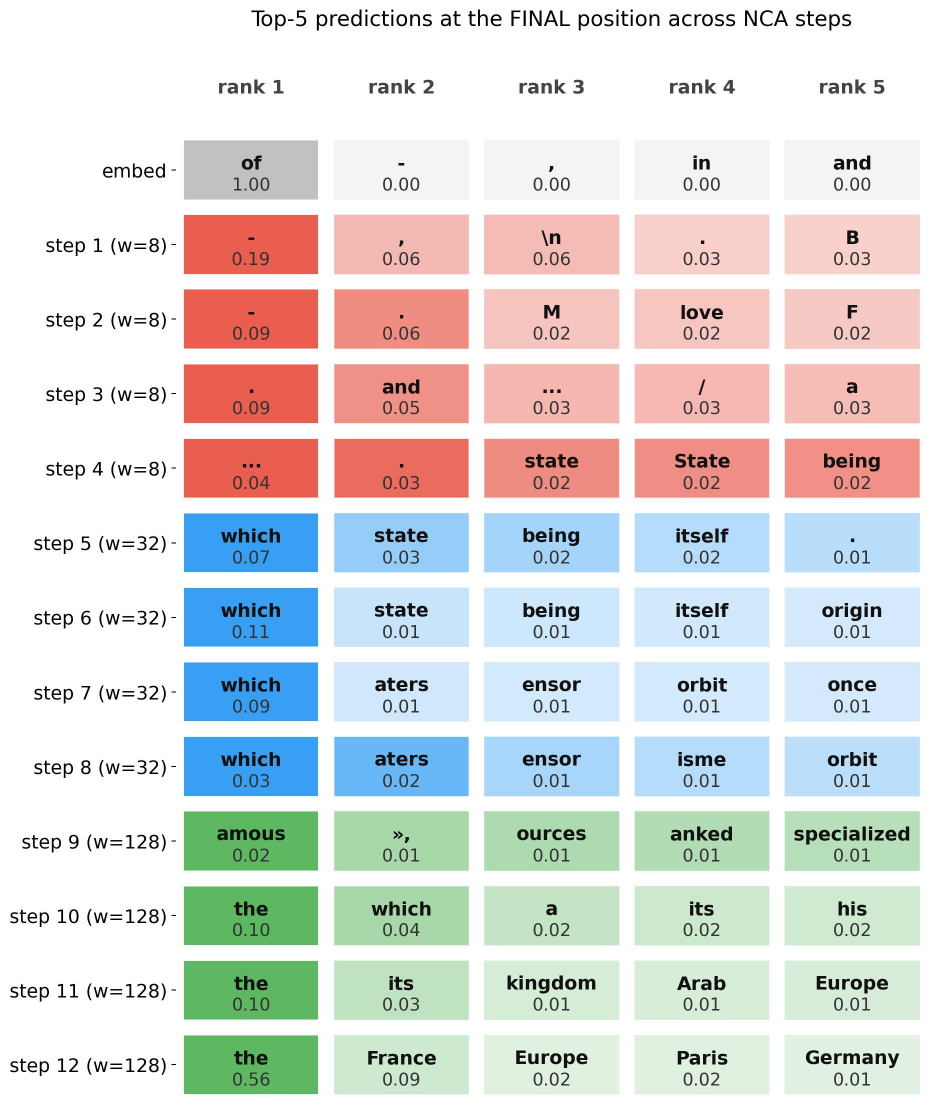}
\caption{Top-5 predicted tokens at the final position across the embedding state and 12 NCA steps. The correct continuation appears only at the final step in this illustrative example.}
\label{fig:topk_final}
\end{figure}

Figure~\ref{fig:topk_final} zooms into the final position of one illustrative prompt and tracks the top-5 predicted tokens across all snapshots. The correct factual continuation only appears in the top-5 at the final step. This example should be read as an illustration of the receptive-field mechanism, not as a general factual-retrieval result. A stronger claim about factual association would require a probe or benchmark across many prompts.

\subsection{Summary}
\label{app:interp_summary}

Across CKA (\S\ref{app:cka}), semantic-clustering similarity (\S\ref{app:similarity}), GRU gates (\S\ref{app:gate_decomp}), within-stage attention (\S\ref{app:attn_progression}), linear probes (\S\ref{app:probing}), hidden-state dynamics (\S\ref{app:hidden_dynamics}), and logit-lens analyses (\S\ref{app:logit_lens}--\ref{app:topk_final}), the stages are clearly not redundant: they produce distinct internal regimes, stage-boundary rewrites, and qualitatively different attention patterns under identical weights. The probe and similarity analyses further show that early stages can be representationally informative (Stage 2 probing peak at $81.7\%$; Paris--France similarity rising to $0.73$ by the start of Stage 3) without contributing measurable next-token loss reduction. We deliberately avoid overinterpreting these internal changes as independently predictive linguistic stages. Combined with the main per-step loss decomposition, the most conservative interpretation is that Hier-TextNCA performs representation preparation in early stages and next-token decoding in the final wide-context stage.

\section{Test-Time Iteration-Count Control: Full Sweep}
\label{app:ttc_full}

\S\ref{sec:ttc} reports the headline test-time iteration-count behaviour. This appendix provides the full inference-time $T_s'$ sweep, the deterministic baseline that motivated stochastic-iteration training, and the mechanistic explanation of why the deterministic flagship has no usable inference-time knob.

\subsection{Why the deterministic flagship has no inference-time knob}
\label{app:ttc_deterministic}

We trained Hier-TextNCA with $T_s = 4$ iterations per stage (12 total NCA steps). A natural question is whether perplexity improves by running more iterations at inference --- a free test-time iteration knob that would be unique to iterative architectures.

\medskip
\noindent
\refstepcounter{figure}\label{fig:ttc_deterministic}%
\begin{center}
\includegraphics[width=\columnwidth]{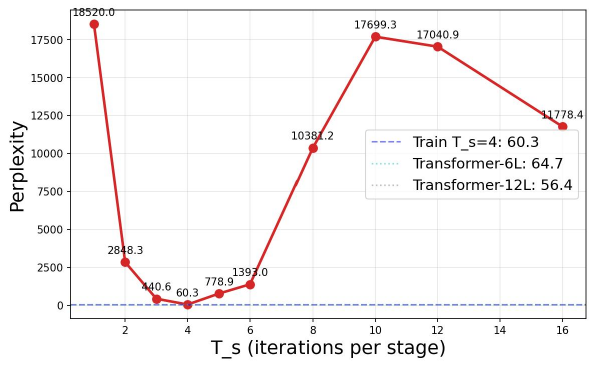}
\end{center}
{\small\noindent\textbf{Figure \arabic{figure}:} WikiText-103 PPL (log scale) for the deterministic Hier-TextNCA flagship evaluated at different inference $T_s'$ (total NCA steps $= 3T_s'$). PPL is U-shaped with the minimum at the training horizon ($T_s'{=}4$, 12 steps).}
\medskip

Figure~\ref{fig:ttc_deterministic} shows the result is negative: perplexity is sharply U-shaped, bottoming at $T_s' = 4$ (PPL 60.3) and exceeding 10{,}000 at $T_s' \in \{1, 8, 10, 12\}$. Two mechanisms explain this.

\paragraph{Learned step embeddings break extrapolation.}
Each of the 12 iterations has its own learned additive embedding $\mathbf{s}_t$. Values of $t$ outside $[0, T_s - 1]$ within a stage are never seen during training, so evaluating with $T_s' > 4$ feeds the perception kernel an out-of-distribution conditioning signal.

\paragraph{No fixed point at the trained horizon.}
As shown in \S\ref{app:hidden_dynamics}, $\|\Delta \mathbf{h}\| \approx 4$ even at step 12. Iterating beyond 12 compounds this drift rather than converging, so the hidden states leave the region the LM head is calibrated to decode.

\subsection{Stochastic-$T_s$ training: positive result with substantial cost}
\label{app:stochastic_t}

Motivated by the deterministic negative result, we train a variant (Hier-StochasticT) with two modifications: (i) sinusoidal step embeddings (deterministic in $t$, so $T_s' > T_s$ at inference is no longer out-of-distribution conditioning); and (ii) per-batch stochastic sampling $T_s \sim \mathrm{Uniform}\{2, 4, 6\}$ during training, so the shared block is explicitly trained to be useful at multiple iteration counts. All other architecture and hyperparameters are identical to the flagship.

\medskip
\noindent
\refstepcounter{figure}\label{fig:ttc_valley}%
\begin{center}
\includegraphics[width=\columnwidth]{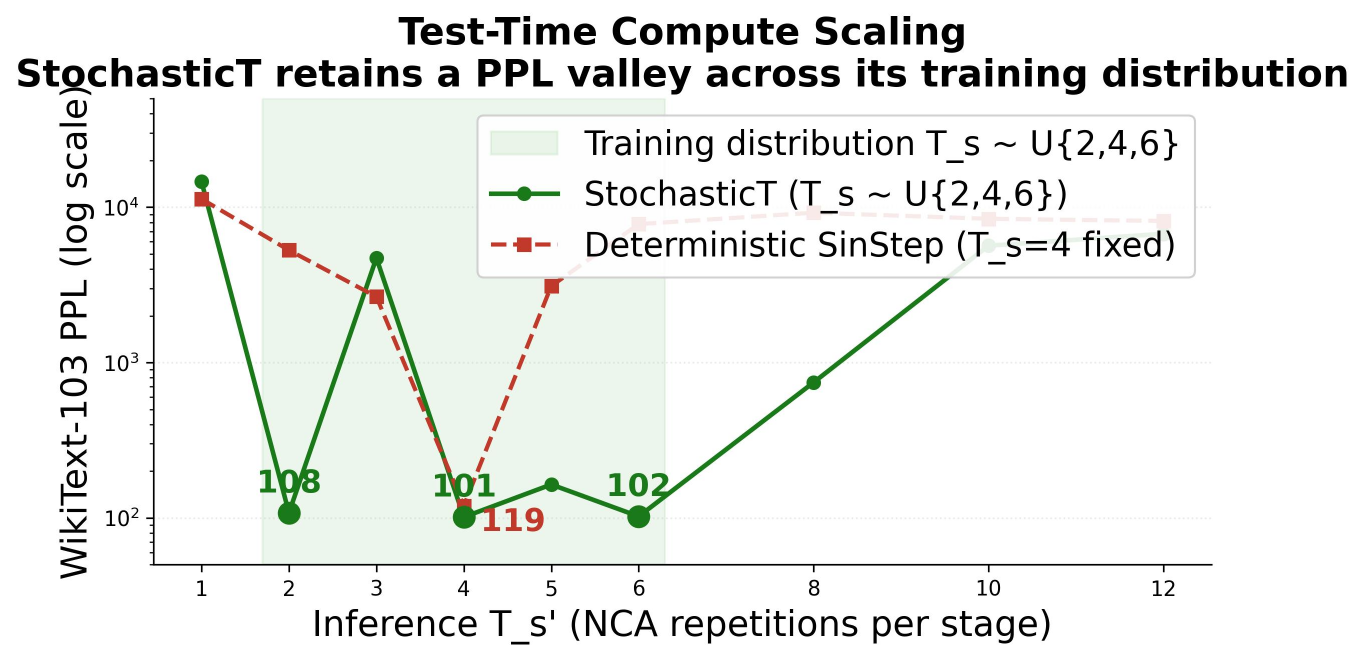}
\end{center}
{\small\noindent\textbf{Figure \arabic{figure}:} Inference-time $T_s'$ sweep, WikiText-103 PPL (log scale). Shaded band: StochasticT training distribution $T_s' \in \{2,4,6\}$. StochasticT (green) keeps a working valley at PPL $101$--$108$; the deterministic SinStep counterpart (red dashed) diverges at any off-distribution $T_s'$.}
\medskip

\begin{table}[h]
\centering
\small
\setlength{\tabcolsep}{2pt}
\begin{tabular}{rrrr}
\toprule
$T_s'$ & Total NCA steps & \textbf{StochasticT} PPL & SinStep (det.) PPL \\
\midrule
1  &  3 & \textit{diverged} & \textit{diverged} \\
\textbf{2}  &  6 & \textbf{108} & \textit{diverged} \\
3  &  9 & \textit{diverged} & \textit{diverged} \\
\textbf{4}  & 12 & \textbf{101} & \textbf{119.5} \\
5  & 15 & 164 & \textit{diverged} \\
\textbf{6}  & 18 & \textbf{102} & \textit{diverged} \\
8  & 24 & 745 & \textit{diverged} \\
10 & 30 & \textit{diverged} & \textit{diverged} \\
12 & 36 & \textit{diverged} & \textit{diverged} \\
\bottomrule
\end{tabular}
\caption{Test-time $T_s'$ sweep on WikiText-103. StochasticT (sinusoidal step + $T_s \sim \mathcal{U}\{2,4,6\}$ training) shows a wide PPL valley spanning the training distribution and partially generalising to $T_s'{=}5$. The deterministic SinStep counterpart (sinusoidal step + fixed $T_s{=}4$ training) diverges at any $T_s' \neq 4$. Bold rows highlight $T_s'$ values inside the StochasticT training distribution. We report values past PPL $\sim 10^3$ as \textit{diverged}: absolute PPLs in this regime are dominated by a near-uniform output distribution and are not meaningfully comparable (raw values: SinStep $\in [2654, 11310]$; StochasticT off-distribution $\in [4714, 14629]$).}
\label{tab:ttc_stochastic}
\end{table}

\paragraph{Headline finding.}
The StochasticT model exhibits a working PPL valley at $T_s' \in \{2, 4, 6\}$ (PPL 101--108) and partial generalisation to $T_s' = 5$ (PPL 164). Outside the training distribution it still degrades, but the in-distribution behaviour is qualitatively analogous to the recent latent-reasoning result of \citet{geiping2025latent} at 3.5B parameters and 800B tokens. The architectural primitives required are simple (sinusoidal step embedding + stochastic-$T_s$ training), and the trade-off is the absolute-PPL cost discussed in \S\ref{sec:ttc}: StochasticT's best PPL (101) is 41 PPL worse than the deterministic flagship (60.3).

\paragraph{What this rules in and rules out.}
In our setting, stochastic-$T_s$ training is the working recipe for an inference-time iteration-count knob. Two alternatives we did not test are (i) removing step conditioning so that the iteration becomes a contraction mapping with a single fixed point, and (ii) adding an explicit convergence criterion at inference. Both remain plausible recipes; we do not claim they would not work, only that they were outside the scope of this study.

\section{FLOP- and Wall-Clock-Matched Comparison}
\label{app:flops_matched}

The main results in Table~\ref{tab:main} report numbers at matched 60k training steps. Because Hier-TextNCA runs $1.9\times$ more FLOPs per forward pass and is $2.7\times$ slower in throughput than Transformer-6L, matching training steps overstates the case for Hier-TextNCA under a typical compute budget. We re-tabulate here under the two stricter protocols, surfaced in \S\ref{sec:main_lm}.

\medskip
\noindent
\refstepcounter{table}\label{tab:flops_matched}%
\begin{center}
\small
\setlength{\tabcolsep}{3pt}
\begin{tabular}{lrrrr}
\toprule
\textbf{Match} & \textbf{Model} & \textbf{Steps} & \textbf{GFLOPs} & \textbf{PPL} \\
\midrule
\multirow{2}{*}{Steps (main)}   & TF-6L         & 60k & 2{,}366 & 52.8 \\
                                & Hier-TextNCA  & 60k & 4{,}531 & 60.3 \\
\midrule
\multirow{2}{*}{Train FLOPs}    & TF-6L         & 60k & 2{,}366 & 52.8 \\
                                & Hier-TextNCA  & 31k & 2{,}341 & 70.6 \\
\midrule
\multirow{2}{*}{Wall-clock}     & TF-6L         & 60k & 2{,}366 & 52.8 \\
                                & Hier-TextNCA  & 23k & 1{,}737 & 80.1 \\
\bottomrule
\end{tabular}
\end{center}
{\small\noindent\textbf{Table \arabic{table}:} Hier-TextNCA vs.\ Transformer-6L at three matching protocols. Under both stricter budgets (FLOPs and wall-clock) the gap to TF-6L widens to 17.8--27.3 PPL. PPL numbers for Hier-TextNCA at non-final step counts are interpolated from the eval-loss curve recorded by the trainer.}
\medskip

\medskip
\noindent
\refstepcounter{figure}\label{fig:pareto}%
\begin{center}
\includegraphics[width=\columnwidth]{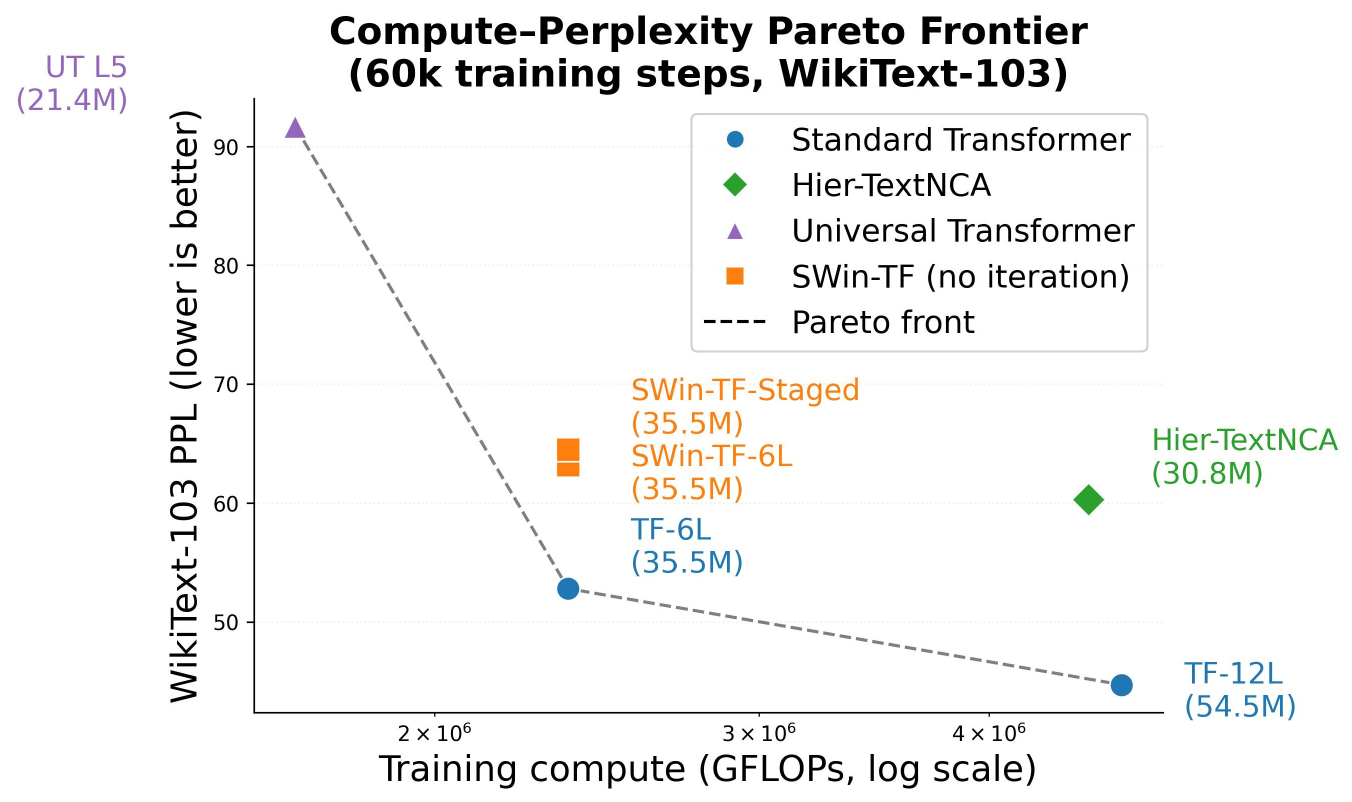}
\end{center}
{\small\noindent\textbf{Figure \arabic{figure}:} Compute--PPL Pareto frontier on WikiText-103 at 60k training steps. Standard transformers (blue) define the frontier; Hier-TextNCA (green) lies $7.5$ PPL behind TF-6L at $1.9\times$ the compute. Dashed line: lower-PPL convex hull. TextNCA is not on the compute--PPL frontier at this scale.}
\medskip

\section{Reproducibility}
\label{app:reproducibility}

We release training code, configuration files (one per row of every results table), and one trained checkpoint per architectural variant under an open-source license. The training entry point is a single \texttt{torchrun} script with one argument selecting the model from a registry; each row of Table~\ref{tab:main}, Table~\ref{tab:ts_full}, and the gating and orchestration ablations corresponds to a registry key documented in the README. Eval scripts produce the JSON \texttt{final\_results.json} files that populate every reported number.

\paragraph{Determinism.}
All runs use a fixed seed of 42 for the trainer; per-rank seeds are 42 plus the rank for data shuffling. We do not enable fully deterministic CUDA kernels because they degrade throughput; rerun-to-rerun variation on a single seed is below 0.05 nats on eval loss in our setup.

\paragraph{Hyperparameters and seeds.}
All hyperparameters are listed in the per-run \texttt{config.json}: per-GPU batch size, gradient accumulation, world size, effective tokens-per-step, learning rate, warmup, total steps, AMP dtype, and tokenizer. Numbers reported as ``mean $\pm$ sd'' are over three random seeds; all others are single-seed and are marked as such in the relevant table caption.

\paragraph{Downstream fine-tuning hyperparameters.}
For the three classification benchmarks (IMDb, AG News, SST-2), each pre-trained checkpoint is fine-tuned with AdamW for $5$ epochs at learning rate $2{\times}10^{-5}$ and batch size $32$, with a mean-pooled classification head added on top of the LM body and trained jointly. For SQuAD~v1.1, we fine-tune for $3$ epochs at learning rate $3{\times}10^{-5}$ and batch size $16$ with the standard span-start / span-end prediction head. All fine-tuning runs reuse sequence length $512$ and the Mistral tokeniser from LM pre-training. Classification numbers in Table~\ref{tab:downstream} are means over the three seeds (42, 1337, 2024); SQuAD is single-seed.

\section{Compute Budget}
\label{app:compute}

All training runs were performed on a single node with two NVIDIA RTX PRO 6000 Blackwell Workstation GPUs (96~GiB each). Per-run compute is summarised in Table~\ref{tab:compute}; numbers are wall-clock and include checkpoint I/O.

\medskip
\noindent
\refstepcounter{table}\label{tab:compute}%
\begin{center}
\small
\setlength{\tabcolsep}{3pt}
\begin{tabular}{lrrr}
\toprule
\textbf{Run} & \textbf{Steps} & \textbf{GPUs} & \textbf{GPU-h} \\
\midrule
Hier-TextNCA $T_s{=}4$ (flagship)        & 60k & 2 & 5.5 \\
Hier-TextNCA $T_s{=}2$ (ablation)        & 60k & 1 & 1.4 \\
Hier-TextNCA $T_s{=}6$ (ablation)        & 60k & 1 & 4.0 \\
Hier-TextNCA $T_s{=}8$ (ablation)        & 60k & 1 & 4.7 \\
Hier-TextNCA gate / kernel abl.\         & 60k & 2 & 5--6 \\
Single-scale TextNCA variants            & 60k & 1--2 & 2--4 \\
Transformer-6L (baseline)                & 60k & 2 & 2.4 \\
Transformer-12L (baseline)               & 60k & 2 & 4.0 \\
Universal Transformer (baseline)         & 60k & 1 & $\sim$1.8 \\
\bottomrule
\end{tabular}
\end{center}
{\small\noindent\textbf{Table \arabic{table}:} Approximate wall-clock GPU-hours per training run; GPU-h is per run for single-design rows and per variant (each) for the two grouped rows (gate / kernel ablations; single-scale variants). The $T_s$ ablations were trained at matched effective batch (65{,}536 tokens/step) on a single GPU. Total project compute including figures-only and interpretability re-runs is approximately 80~GPU-hours.}
\medskip

\paragraph{Inference cost.}
A forward pass of Hier-TextNCA at sequence length 512 takes 75.5~GFLOPs versus 39.4~GFLOPs for Transformer-6L and 78.7~GFLOPs for Transformer-12L. End-to-end throughput at batch 32, sequence length 512, on one Blackwell-6000 is 53k tokens/s for Hier-TextNCA versus 138k tokens/s for Transformer-6L ($2.6\times$ slower). All FLOP counts use standard self-attention plus FFN accounting and are reproducible from the released configs.

\section{Use of AI Assistants}
\label{app:ai_use}

We used AI coding assistants during this project for two purposes: helping write and debug implementation code (model definitions, training and evaluation scripts, figure-generation scripts) and assisting with paper writing (LaTeX formatting, editorial revisions, consistency checks across sections, and table formatting). The ideation, research questions, experimental design, choice of ablations, the architectural design of \textsc{TextNCA}, and the interpretation of all empirical results are the authors' own. All AI suggestions were reviewed and verified by the authors before being committed to the codebase or to the paper.

\end{document}